\documentclass[12pt]{article}
\usepackage{amsmath}
\usepackage{graphicx}
\usepackage{enumerate}
\usepackage{url} % not crucial - just used below for the URL 
\usepackage[T1]{fontenc}

\usepackage{amsmath,amssymb,amsthm,mathrsfs}
\usepackage{mathrsfs}
\usepackage{dsfont}
\usepackage[english]{babel}
\usepackage{graphicx,color}
\usepackage{enumitem}
\usepackage[colorlinks]{hyperref}
\usepackage{bm}

\hypersetup{
linkcolor=blue,
citecolor=blue,
}
\usepackage{siunitx} %

\graphicspath{{Figures/}}
\usepackage{subcaption}

\RequirePackage{algorithm}
\RequirePackage{algorithmic}
\usepackage{natbib}
\usepackage{microtype}

\usepackage{booktabs} 
\usepackage{hyperref}
\usepackage{mathtools}
\mathtoolsset{showonlyrefs}

\usepackage{amsmath,amsfonts}
\usepackage{scalerel,amssymb}
\usepackage{amsthm}
\newcommand{\R}{{\rm I\!R}}

\newcommand{\W}{\mathcal{W}}

\newcommand{\bmu}{\bm{\mu}}

\newcommand{\X}{\mathbf{X}}

\newcommand{\tbF}{\widetilde{\bm F}}

\DeclareMathOperator*{\argmin}{arg\,min}
\DeclareMathOperator{\Log}{Log}
\DeclareMathOperator{\Exp}{Exp}
\DeclareMathOperator{\Tan}{Tan}

\newtheorem{assump}{Assumption}

\newcommand{\WR}{\W_2(\R)}

\theoremstyle{plain}
\newtheorem{theorem}{Theorem}[section]

\newtheorem{prop}[theorem]{Proposition}
\newtheorem{definition}{Definition}[section]

\theoremstyle{remark}

\usepackage{mathabx}

\numberwithin{equation}{section} 

\newcommand{\CR}{\textcolor{black} } %  Commentaires Jeremie
\newcommand{\CB}{\textcolor{black} } %  modifications

\newcommand{\blind}{1}

\begin{document}

\def\spacingset#1{\renewcommand{\baselinestretch}%
{#1}\small\normalsize} \spacingset{1}

%%%%%%%%%%%%%%%%%%%%%%%%%%%%%%%%%%%%%%%%%%%%%%%%%%%%%%%%%%%%%%%%%%%%%%%%%%%%%%

\if1\blind
{
  \title{\bf Wasserstein auto-regressive models for modeling multivariate distributional time series}
  \author{Yiye Jiang \thanks{ This author is supported by MIAI@Grenoble Alpes, (ANR-19-P3IA-0003). Part of the work of this author was conducted while she was preparing her PhD at Université de Bordeaux. 
    %The authors gratefully acknowledge \textit{please remember to list all relevant funding sources in the unblinded version}
    }\hspace{.2cm}\\
    Universit\'e Grenoble Alpes, CNRS, Inria, Grenoble INP, LJK. \\
    and \\
    J\'{e}r\'{e}mie Bigot \\
    Institut de Math\'ematiques de Bordeaux, Universit\'e de Bordeaux}
  \maketitle
} \fi

\if0\blind
{
  \bigskip
  \bigskip
  \bigskip
  \begin{center}
    {\LARGE\bf Wasserstein auto-regressive models for modeling multivariate distributional time series}
\end{center}
  \medskip
} \fi

\bigskip
\begin{abstract}
This paper is focused on the statistical analysis of data
consisting of a collection of multiple series of probability measures that are indexed by distinct time instants and supported over a bounded interval of the real line. By modeling these time-dependent probability measures as random objects in the Wasserstein space, we propose a new auto-regressive model for the statistical analysis of multivariate distributional time series.  Using the theory of iterated random function systems, results on the   second order stationarity of the solution of such a model are provided. We also propose a consistent estimator for the auto-regressive coefficients of this model. Due to the simplex constraints that we impose on the model coefficients, the proposed estimator that is learned under these constraints, naturally has a sparse structure. The sparsity allows the application of the proposed model in learning a graph of temporal dependency from   multivariate distributional time series. We explore the numerical performances of our estimation procedure using simulated data. To shed some light on the benefits of our approach for real data analysis, we also apply this methodology to two data sets, respectively made of observations from age distribution in different countries and those from the bike sharing network in Paris.
\end{abstract}

\noindent%
\vfill

\noindent%
\vfill

\newpage
\spacingset{1.9} % DON'T change the spacing!

\section{Introduction}

\begin{figure}
    \centering
    \hspace{0.2in}
    \includegraphics[width=0.45\linewidth]{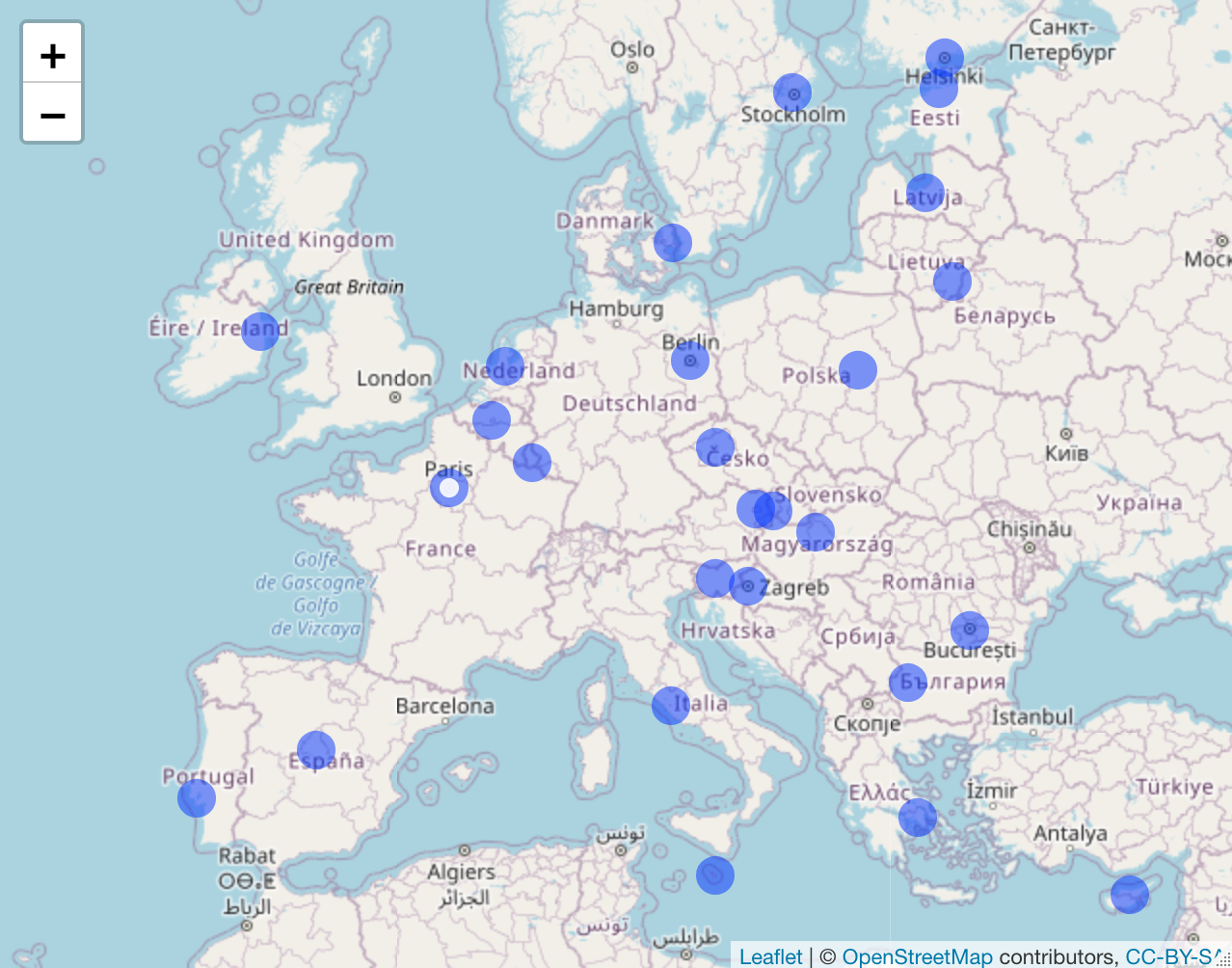}\\
    \vspace{0.1in}
    \includegraphics[width=0.48\linewidth]{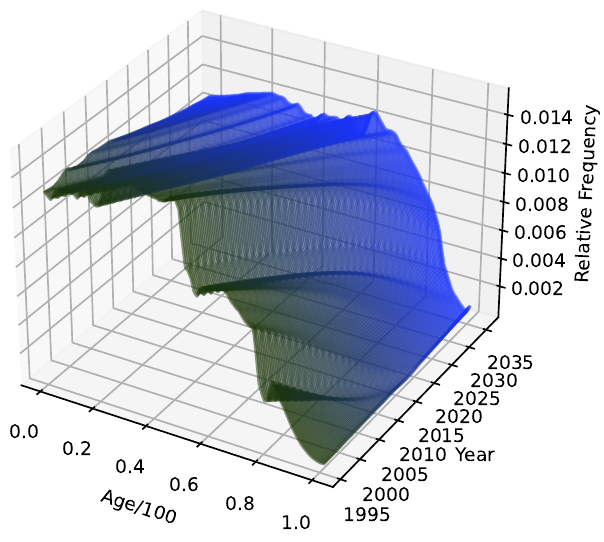}
 \includegraphics[width=0.49\linewidth]{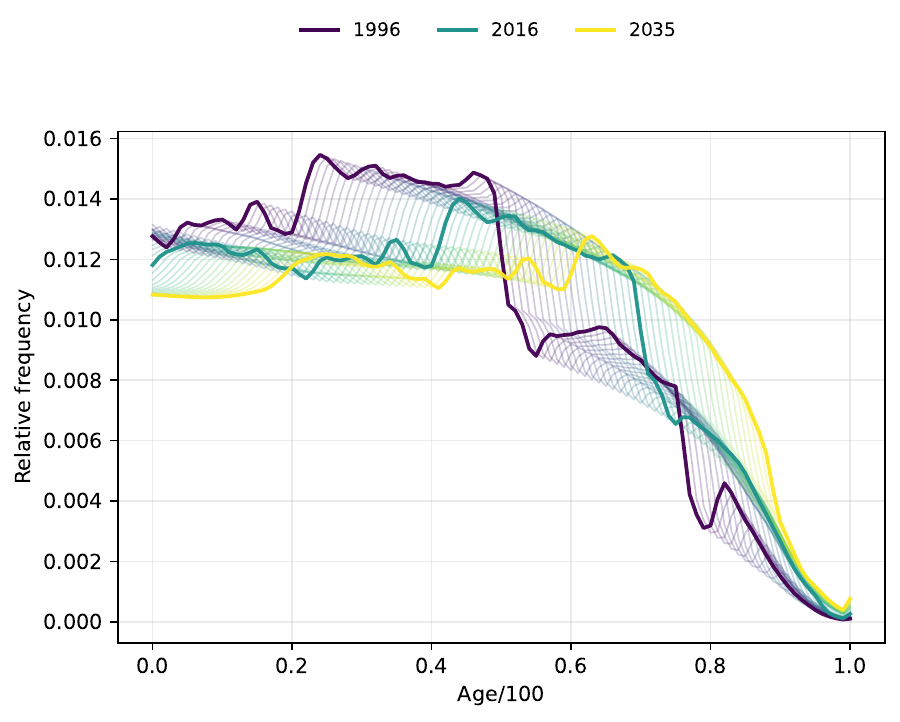}
    \caption{\textit{Annual records of age distributions of EU countries.} On the top are $27$ countries in the European union. A sequence of age distribution is recorded at each country over years. For example, at the bottom we illustrate the sequence of France, where one distribution supported over $[0,1]$ is observed at each year. On the lower left, we visualize the resulting univariate distributional time series with a surface in the coordinate system of Age $\times$ Year $\times$ Relative frequency. The raw data in this plot consist in $40$ annual distributions. We complete them with interpolated samples to draw the surface. On the lower right, we show the projection of the raw time series onto the Age $\times$ Relative frequency plane. We can see that the population is aging along time. }
    \label{fig-intro: distributional}
\end{figure}

Distributional time series is a recent research field that deals with observations that can be modeled as sequences of time-dependent probability distributions. Such distributional time series are ubiquitous in many scientific fields. A pertinent example is the analysis of sequences of the indicator distributions supported over age intervals, such as mortality and fertility \citep{mazzuco2015fitting, shang2020forecasting}, over calendar years in demographic studies. Other examples include daily stock return distributions from financial time series \citep{kokoszka2019forecasting, zhang2021wasserstein}, the distributions of correlations between pairs of voxels within brain regions \citep{petersen2016functional}. Figure \ref{fig-intro: distributional} displays an illustrative example, a dataset made of time series of age distributions for countries in the European Union.

Probability distributions can be characterized by functions, such as densities, quantile functions, and cumulative distribution functions. Therefore, to analyze  distributional time series, one may study one of its functional representations using tools from functional time series analysis \citep{bosq2000linear}. However, due to their nonlinear constraints, such as monotonicity and positivity, the representing functions of distributions do not constitute linear spaces. Consequently, basic notions in functional time series models, such as additivity and scalar multiplication, do not adapt, in a straightforward manner for random elements not belonging to a Hilbert space. An existing approach is mapping the densities of distributions to unconstrained functions in a Hilbert space by the log quantile density (LQD) transformation \citep{petersen2016functional}, and then apply the standard tools. However LQD does not take into account the geometry of the space of probability distributions.  Recent approaches naturally turn to the development of new time series models tailored for distributions in some space of probability measures. Among others, a widely considered space is the one of Wasserstein \citep{chen2019lqd, horvath2021monitoring, layapanaretos}.

An important family of time series model is auto-regressive (AR) models. The first works extending  standard AR models in Wasserstein space are \cite{chen2019lqd, layapanaretos, chen2021wasserstein,zhang2021wasserstein,zhu2023autoregressive}. They consider in particular the observation of a single series of univariate distributions. These models allow to analyze the dependency between probability distributions observed on consecutive time points, for instance, the age distributions of a population recorded in two successive years. 

\CB{However, to the best of our knowledge, no work has yet developed AR models dedicated to the multiple distributional time series, such as those displayed in Figure \ref{fig-intro: distributional}. The resulting models will allow richer analysis, as they enable the study of both within-series and cross-series dependencies. For example, one may investigate the dependency of age distributions of different countries, which indicates how changes in the age distribution of one country propagate to other countries. This is the motivation of this paper. More specifically, we consider the extension of vector AR (VAR) model of order $1$ as defined in Equation \eqref{eq:var0} for a collection of time series of univariate distributions, based on the geometry of Wasserstein space.}
\begin{definition}
Given a multivariate time series $\bm x_t^i \in \R, \, t \in \mathbb{Z}, \, i = 1, \dots, N$, such that the expectation is time invariant for each component $i$  that is $u^i = \mathbb{E}(\bm x_t^i), \, t \in \mathbb{Z}$. Then the VAR model of order $1$ (not including an intercept term) writes as \cite{helmut2005new}
\begin{equation}\label{eq:var0}
 \bm x_t^i = u_i + \sum_{j=1}^N A_{ij} (\bm x_{t-1}^j - u_j) + \bm \epsilon_t^i,
\end{equation}
where $\bm \epsilon_t^i$ is a white noise.     
\end{definition}
The difficulty in extension is, as mentioned previously, the extension of addition and scalar multiplication into a space, which is not linear.

%make better understand the utility of the porposed models 

%As for the regression models in Wasserstein space, similarly to the situation of auto-regressive models, almost all the existing tools consider only one predictor distribution (see for example \cite{ghodrati2022distribution}). One recent work devised for multiple predictor distributions is \cite{zhu2023geodesic}. They propose to rely only on optimal transport maps and a sequence of pushforwards to construct the regression. Such intrinsic method in the multivariate case will lead to an ordering problem, because pushing forward with the same set of transport maps but in different orders will give different distributions. To avoid such challenge, we propose to rely on the tangent space, where the addition is well defined between tangent vectors (essentially transport maps). 

\subsection{Main contributions}

In this paper, we introduce a new AR model for multivariate distributional time series. The properties of this model can be analyzed thanks to the theory of iterated random function systems \cite{wu2004limit,zhu2023autoregressive}. Moreover, due to the geometric constraints inherited from the Wasserstein space that are imposed on the AR coefficients of the model, the estimator that we propose naturally has a sparse structure. In particular, this allows the application of our approach to  learning a sparse graph of temporal dependency from   multivariate distributional time series. The numerical performances of our approach are then illustrated with simulated data. Finally, the methodology is applied to the real data set of age distribution displayed in Figure \ref{fig-intro: distributional}, in addition, to another real data set based on the bike-sharing network in Paris.
 
\subsection{Organization of the paper}
In Section \ref{sec:back}, we provide the background on the geometry of the Wasserstein space. In Section \ref{sec: uni AR wass}, we review a univariate AR model in literature that inspires directly our models. In Section \ref{sec: wmar}, we derive the proposed Wasserstein multivariate AR models. We also study its second order stationarity.  A consistent estimator of the AR coefficients is constructed in Section \ref{sec: estimation}. Finally, numerical experiments using simulated and real data are carried out in Section \ref{sec: num_exp} to analyze the finite sample  properties of the estimator,  and to illustrate its application in graph learning from multivariate distributional time series.

\subsection{Publicly available source code}

For the sake of reproducible research, Python code available at %\textit{the github of the first author}
\begin{center}
\url{https://github.com/yiyej/Wasserstein_Multivariate_Autoregressive_Model}
\end{center}
implements the proposed estimators and the experiments
carried out in this paper. %\textit{To respect the double anonymous policy, the explicit link is not given in the current version for peer review. }

\section{Background on the Wasserstein space} \label{sec:back}
%\color{blue}
In this paper, we model the observations of distribution as probability measures in a Wasserstein space. We consider the specific space, $\W_2(\R)$ due to its analytical facility such as the explicit formula of its optimal transport maps. In the following we only introduce the necessary notions in $\W_2(\R)$ to our development, which can be all found in \cite{panaretos2020invitation} with a comprehensive introduction on the Wasserstein space. In the following, we use distribution and probability measure interchangeably.

We introduce first the definition of $\W_2(\R)$. Let $\mathcal{P}(\R)$ be the set of all the probability measures over measurable space $(\R, \mathcal{B}(\R))$ and $\mathcal{B}(\R)$
the associated $\sigma$-algebra made of Borel subsets of $\R$.  Then we consider the subset of $\mathcal{P}(\R)$:
\begin{equation}
 \mathcal{P}_2(\R) = \left\{ \mu \in \mathcal{P}(\R) \Big| \int_\R x^2 d \mu(x) < \infty \right\},
\end{equation}
which are all probability measures over $\R$ with finite second moment. Equipped this set with $\mathcal{L}^2$ Wasserstein distance:
\begin{equation}\label{eq: wass_dist}
    d_W(\mu, \nu) = \left( \int_{0}^{1} \left[F_{\mu}^{-1}(p) - F_{\nu}^{-1}(p)\right]^2 \,dp\right)^{1/2}, \quad \mu, \nu \in \mathcal{P}_2(\R), 
\end{equation}
where $F_{\mu}^{-1}, F_{\nu}^{-1}$ are respectively quantile functions\footnote{We recall the definition of quantile functions in the supplemental materials.} of $\mu$ and $\nu$, we obtain the Wasserstein space $\W_2(\R)$. The $\mathcal{L}^2$ Wasserstein distance is simply the standard $\mathcal{L}^2$ distance between two quantile functions.  

%It is well known that $\W$ is a complete and separable metric space (see e.g.\ \cite{villani2021topics} for a detailed course  on optimal transport theory and \cite{panaretos2020invitation} for an introduction to the topic of statistical analysis in the Wasserstein space). 

\begin{figure}
    \centering
    \includegraphics[width=0.7\linewidth]{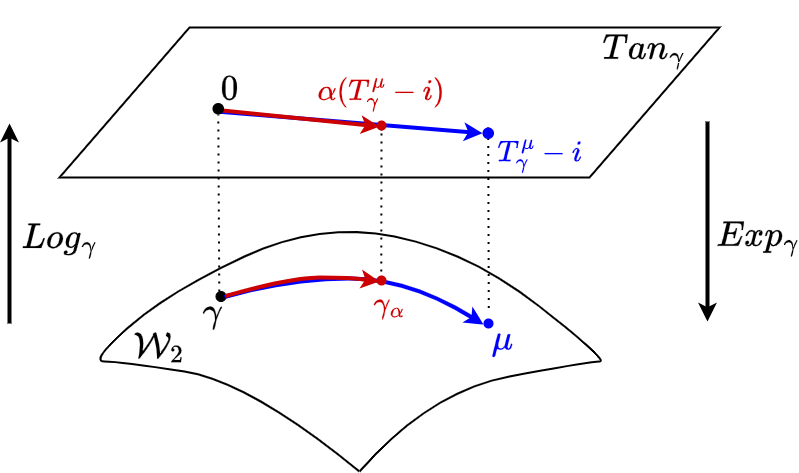}
    \vskip -0.25in
    \caption{Geodesic in $\W_2(\R)$.}
    \label{fig: geodesic_Wass}
\end{figure}

One of the key steps in the extension is interpreting the subtractions in the standard VAR model \eqref{eq:var0}, $\bm x_{t-1}^j - u_j$, by geodesic. Since the notion of geodesic exists also in the Wasserstein space, we can extend the subtractions for distributions. In the following we define the geodesic. Given two measures $\gamma, \mu \in \W_2(\R)$, the geodesic is constructed in three steps. First, a Hilbert space with nice properties is constructed, along with two mappings so as to embed the Wasserstein space into the space of functions. Second, $\gamma, \mu$ are mapped into the Hilbert space and a geodesic is defined there which is simply a line segment. Third, one maps back the segment into the space of distributions, the image is the geodesic. The Hilbert space is named tangent space, and the two maps are exponential and logarithmic maps. We first present the tangent space. 
\begin{definition}
Every $\gamma \in \W_2(\R)$ that possesses a continuous cumulative distribution function (cdf) $F_{\gamma}$ allows a tangent space:
\begin{equation}\label{eq: tengent space}
    \Tan_{\gamma} = \overline{\{t(T_\gamma^\mu - id): \mu \in \W_2, \; t > 0\}}^{\mathcal{L}^2_{\gamma}},
\end{equation}
where $T_\gamma^\mu = F_{\mu}^{-1}\circ F_{\gamma}$ is the optimal transport map, that pushes $\gamma$ forward to $\mu$.  
$\Tan_{\gamma}$ is endowed with the inner product $\langle\cdot,\cdot \rangle_{\gamma}$ defined by 
$$\langle f,g \rangle_{\gamma} := \int_{\R} f(x)g(x) \,d\gamma(x), \; f, g \in \mathcal{L}^2_{\gamma}(\R),$$
and the induced norm $\|\cdot\|_{\gamma}$.    
\end{definition}
An optimal transport map is a function that characterizes the difference between two probability measures, which can be understood as the counterpart of vector of subtraction $X-Y, \, X, Y \in \R^p$, in the context of distributions. 
%It also indicates how one can transform a measure to another measure in an optimal way.
It is unique for the two probability measures in comparison. 
When fixing $\gamma$, we can associate any measure $\mu$ in $\W_2(\R)$ to a function, given by their optimal transport map, $T_\gamma^\mu$. This is the starting point of tangent space, and the mappings. For the technical facility that the reference measure $\gamma$ can be mapped to zero, we associate $\mu$ to $T_\gamma^\mu - id$ instead of $T_\gamma^\mu$, thus when $\mu = \gamma$, $T_\gamma^\mu - id = T_\gamma^\gamma - id = id-id=0$. The argument implies the mappings as follows. 
\begin{definition}\label{def: log}
The logarithmic map $\Log_{\gamma}: \W_2(\R) \rightarrow \Tan_{\gamma}$ is defined as
\begin{equation}
    \Log_{\gamma} \mu = T_\gamma^\mu - id. 
\end{equation}
\end{definition}
\begin{definition}\label{def: exp}
The exponential map $\Exp_{\gamma}: \Tan_{\gamma} \rightarrow \W_2(\R)$ is defined as
\begin{equation}\label{eq: exp}
    \Exp_{\gamma}g = (g + id)\#\gamma, 
\end{equation}
where for any measurable function $T: \R \rightarrow \R$ and $\mu \in \W_2(\R)$, $T\#\mu$ is the pushforward measure on $\R$ defined as $T\#\mu (A) = \mu(\{x \in \R: T(x) \in A\})$, for any set $A\in\mathcal{B}(\R)$. 
\end{definition}
The logarithmic map maps a measure to a function in the tangent space with $\gamma$ as reference measure. The exponential map does the reverse with the operation pushforward. Pushforward transforms a measure to another through applying the difference characterized by a function. Now, we present the geodesic. 
\begin{definition}
Let $\gamma \in \W_2(\R)$ with a continuous cdf, and $\mu \in \W_2(\R)$. The geodesic (McCann's interpolant) between $\gamma$ and $\mu$ is defined as
\begin{equation}
\Exp_{\gamma} [\alpha(T_\gamma^\mu - id)], \; \alpha: 0 \rightarrow 1. 
\end{equation}
\end{definition}
Given that $T_\gamma^\mu - id$ is the function associated to $\mu$ in the tangent space, and $\gamma$ is mapped to 0, $\alpha(T_\gamma^\mu - id)$ is the line segment connecting $0$ and $T_\gamma^\mu - id$. The rest of the definition just maps this "shortest path" back to the Wasserstein space. Furthermore, using the function $T_\gamma^\mu - id$, one can compute any weighted interpolant in a similar intuitive way, as shown in the following proposition.
\begin{prop}\label{prop: interpolant}
Define the probability measure $\gamma_\alpha, \, \alpha \in [0,1]$ by
\begin{equation}
\gamma_\alpha = \Exp_{\gamma} [\alpha(T_\gamma^\mu - id)].
\end{equation}
Then $d_W(\gamma, \gamma_{\alpha}) = \alpha d_W(\gamma, \mu),$ where $d_W$ is the Wasserstein distance of $\W_2(\R)$.
\end{prop}
For example, if $\alpha = 0.5$, one can compute the midpoint between $\gamma$ and $\mu$ by taking the image of the point “halfway” between $0$ and $T_\gamma^\mu - id$ in the tangent space. We summarize all the involved notions so far in Figure \ref{fig: geodesic_Wass}.
%This property will be used to extend the interpolant, $u_i + A_{ij}(\bm x_t^j - u_j)$ of VAR model \eqref{eq:var0}.

%Note that the tangent space change each time, which is not a pb for deriving a univariate AR model in literature, but this cause difficulty in the multivariate setting, which is our case. 

Lastly, we need a notion of mean in the Wasserstein space so as to replace the standard expectation $u_i$ in the VAR models. An appropriate notion is Fréchet mean which is a generalized expectation for metric space. The definition in the context of $\W_2(\R)$ is given as below. 
\begin{definition}\label{def: frechet mean}
Let $\bmu$ be a random measure\footnote{Definition of random measure see the appendices.} from a probability space $(\Omega, \mathcal{F}, \mathbb{P})$ to $\W_2(\R)$. Assume that $\bmu$ is square integrable, namely $\mathbb{E}d^2_W(\bmu, \nu) < \infty$ for some (thus for all) $\nu \in \W_2(\R)$. Then the population Fr\'echet mean of $\bmu$, denoted by $\mathbb{E}_\oplus\bm\mu$, is defined as the unique minimizer of 
\begin{equation}
    \min_{\nu \in \W_2(\R)} \mathbb{E}\left[d^2_W(\bmu, \nu)\right]. 
\end{equation}
Furthermore, $\mathbb{E}_\oplus\bm\mu$ admits an explicit form:
\begin{equation}
    F_{\mathbb{E}_\oplus\bm\mu}^{-1}(p) = \mathbb{E}\left[ \bm F_{\bmu}^{-1}(p) \right], \; p \in (0, 1),
\end{equation}
where $F_{\mathbb{E}_\oplus\bm\mu}^{-1}$ and $\bm F_{\bmu}^{-1}$ are respectively quantile functions of $\mathbb{E}_\oplus\bm\mu$ and $\bmu$. 
\end{definition}
Recall that the usual mean in $\mathbb R^p$ can be written as  
\[
\mathbb{E} \mathbf X = \arg\min_{U \in \mathbb R^p} \; \mathbb{E}\big\|\mathbf X - U\big\|_{2}^2.
\]  
Therefore, the Fréchet mean is based on the same metric‐minimization viewpoint: it generalizes the classical mean simply by replacing the $l_2$ metric with a general metric.

In what follows, we use bold notation to distinguish random quantities from constant (that is non-random) ones.

\section{Wasserstein univariate AR models}\label{sec: uni AR wass}
\begin{figure}
    \centering
    \includegraphics[width=0.5\linewidth]{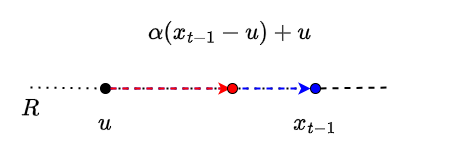}
    \caption{Geometric interpretation of standard univariate $AR$ model \eqref{eq:ar}.}
    \label{fig: geo interp ar}
\end{figure}

In this section, we review the univariate AR models proposed in \cite{zhu2023autoregressive}, which inspires directly our models. Consider the univariate AR models in $\R$:
\begin{equation}\label{eq:ar}
 \bm x_t = u + \alpha (\bm x_{t-1} - u) + \bm \epsilon_t,
\end{equation}
where $\bm x_t, u, \epsilon_t \in \R$, with $\mathbb{E}\bm x_t = u, \, t \in \mathbb{Z}$, $\mathbb{E}\bm \epsilon_t = 0, \, t \in \mathbb{Z}$, and $\bm\epsilon_t$ is independent of $\bm x_{t^\prime}, \, t^\prime < t$. To extend the model for a series of univariate distributions $\bm \mu_t \in \W_2(\R)$, one first assumes that the stationary mean exists, that is, $\mathbb{E}_\oplus\bm \mu_t = \mu_\oplus, \, t \in \mathbb{Z}$. \cite{zhu2023autoregressive} then propose to interpret the subtraction $\bm x_{t-1} - u$ by the geodesic between the observation at time $t-1$, $\bm x_{t-1}$, and the mean of the series $u$. Thus, $u + \alpha (\bm x_{t-1} - u)$ is interpreted as the interpolant of $\bm x_{t-1}$ and $u$ on the geodesic with weight $\alpha$, as illustrated in Figure \ref{fig: geo interp ar}. The notion can be reproduced in the Wasserstein space with Proposition \ref{prop: interpolant}. This leads to the Wasserstein univariate AR:
\begin{equation}\label{eq: uni AR wass}
 \bm \mu_t  = \bm \Gamma_{\bm\epsilon_t}\circ \Exp_{\mu_{\oplus}}\left\{\alpha(\bm T^{t-1}_{\oplus}-id)\right\},
\end{equation}
where $\bm T^{t-1}_{\oplus} = \bm F_{\bm\mu_{t-1}}^{-1}\circ F_{\mu_\oplus}$ is the optimal transport map between the observed distribution at time $t-1$, $\bm\mu_{t-1}$, and the mean distribution of the series  $\mu_\oplus$, and $\bm \Gamma_{\bm\epsilon_t}: \W_2(\R) \rightarrow \W_2(\R)$ is some random map that represents the noise, that we will detail later when developing our model. 

\section{Wasserstein multivariate AR Models}\label{sec: wmar}
\begin{figure}
    \centering
    \includegraphics[width=0.8\linewidth]{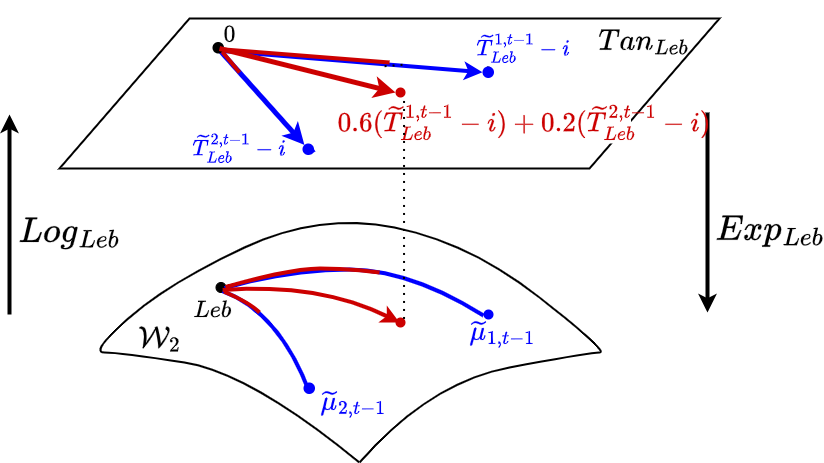}
    \caption{Geometric interpretation of Wasserstein multivariate AR model \eqref{eq: multi_AR_Wass}. The figure corresponds to a simple regression formula with $N=2$. The red point represents the value of $\Exp_{Leb}\left\{ 0.6(\widetilde{\bm T}^{1,t-1}_{Leb}-id) + 0.2(\widetilde{\bm T}^{2,t-1}_{Leb}-id)\right\}$, which is closer to $\widetilde{\bm \mu}_{1,t-1}$ due to a larger regression coefficient. }
    \label{fig: multi_ar_wass_interp}
\end{figure}

In this section, we present the proposed model. The goal is extending the multivariate AR models, namely, VAR models that we recall again in the following equation:
\begin{equation}\label{eq:var0}
 \bm x_t^i = u_i + \sum_{j=1}^N A_{ij} (\bm x_{t-1}^j - u_j) + \bm \epsilon_t^i,
\end{equation}
where $\bm x_t^i \in \R, \, i = 1, \ldots, N, t \in \mathbb{Z}$, $\mathbb{E}\bm x_t^i = u_i,$ $\mathbb{E}\bm \epsilon_t^i = 0$, and $\bm \epsilon_t^i$ are independent of $\bm x_{t'}^i, t' < t$. To extend the VAR model for $N$ series of univariate distributions $\bm \mu_t^i \in \W_2(\R)$, one first assumes that the stationary mean exists for each series as stated in Assumption \ref{assump: stationary N means}. 
\begin{assump}\label{assump: stationary N means}
$\mathbb{E}_\oplus\bm \mu_t^i = \mu_{i,\oplus}, \, i = 1, ..., N, t \in \mathbb{Z}$.
\end{assump}
We would like to rely on the same idea of extension, which extends the subtraction $\bm x_{t-1} - u$ to the geodesic between $\bm\mu_{t-1}$ and $\mu_\oplus$, namely, $\Exp_{\mu_{\oplus}}\left\{\alpha(\bm T^{t-1}_{\oplus}-id)\right\}$. However, if we replace the $N$ subtractions $\bm x_{t-1}^j - u_j, j = 1, ..., N$, analogously with the  $N$ geodesics between $\bm\mu_{t-1}^j$ and $\mu_{j,\oplus}$, $\Exp_{\mu_{j,\oplus}}\left\{A_{ij}(\bm T^{j,t-1}_{j,\oplus}-id)\right\}$, it is not clear how to proceed with the addition. Because the calculations of geodesics in the Wasserstein space are defined from tangent spaces. If we wish to sum the geodesics $\Exp_{\mu_{j,\oplus}}\left\{A_{ij}(\bm T^{j,t-1}_{j,\oplus}-id)\right\}$, we need to sum them firstly in a tangent space. However, they are based on $N$ different tangent spaces, thus $\Tan_{\mu_{j,\oplus}}$, their functions $A_{ij}(\bm T^{j,t-1}_{j,\oplus}-id)$ can not be calculated in one equation. To contour this problem, we propose to ``center'' the series of distributions, so that all component series share the same mean, namely, $\mu_{i,\oplus} \equiv \mu_{\oplus}$. We will furthermore make $\mu_{\oplus} = \mathcal{U}(0,1)$, that is the uniform distribution over $(0,1)$, denoted by $Leb$. The proposed centering method is developed in Section \ref{sec: centering}. Thanks to the unique mean, we can now define the addition of geodesics from $\Tan_{Leb}$. This brings to the proposed model:
\begin{equation}\label{eq: multi_AR_Wass}
    \widetilde{\bm\mu}_t^i = \bm \Gamma_{\bm\epsilon_t^i}\circ\Exp_{Leb}\left\{\sum_{j=1}^N A_{ij}(\widetilde{\bm T}^{j,t-1}_{Leb}-id)\right\}, \quad i = 1, ..., N, 
\end{equation}
where $\widetilde{\bm T}^{j,t-1}_{Leb} = \widetilde{\bm F}_{{j,t-1}}^{-1} \circ id = \widetilde{\bm F}_{{j,t-1}}^{-1}$ is the optimal transport map between the centered observation $\widetilde{\bm\mu}_{t-1}^j$ and $\mathcal{U}(0,1)$, and $\bm \Gamma_{\bm\epsilon_t}: \W_2(\R) \rightarrow \W_2(\R)$ is a random map representing the noise. We detail $\bm \Gamma_{\bm\epsilon_t^i}$ later in the technical section \ref{sec: identifiability}. An illustration of model interpretation is given in Figure \ref{fig: multi_ar_wass_interp}, where it shows that the regression formula $\Exp_{Leb}\left\{\sum_j A_{ij}(\widetilde{\bm T}^{j,t-1}_{Leb}-id)\right\}$ can still be seen as an interpolant between the mean $Leb$ and the observed distributions (centered) at previous time from different series, $\bm \widetilde{\bm\mu}_{t-1}^j, j = 1, ..., N$, with weights given by $A_{ij}$. In particular, a larger weight pulls the interpolant distribution closer to the observed distribution in the corresponding series. This justifies that the fitted regression coefficients in model \eqref{eq: multi_AR_Wass} can be used to study cross-series similarity.    
Model \ref{eq: multi_AR_Wass} needs assumptions to guarantee the identifiability and second order stationarity as a time series system. We treat them respectively in technical sections \ref{sec: identifiability} and \ref{sec: Existence, uniqueness and stationarity}. 

\subsection{Centering a univariate random distribution}\label{sec: centering}
The idea of centering comes from the classical time series analysis where sometimes it is needed to center a time series $\bm x_t \in \R$ by $\bm x_t - u$, with $u$ the stationary series mean, so that the new series mean becomes $0$. For our model \eqref{eq: multi_AR_Wass}, we wish to center each component series so that the new stationary mean becomes $\mathcal{U}(0,1)$. This means, for series $i$, transforming each of its distributions $\bm\mu_{t}^i, t \in \mathbb{Z}$, with some operation based on the original series mean, $\mu_{i,\oplus}$, so that the transformed distribution $\widetilde{\bm\mu}_{t}^i$ admits that $\mathbb{E}_\oplus\widetilde{\bm\mu}_{t}^i = \mathcal{U}(0,1)$. In the following, we construct the operation via quantile functions. The method applies to any random distribution, not necessarily in a time series. 

\begin{figure}[htbp]
    \centering    \includegraphics[width=0.32\linewidth]{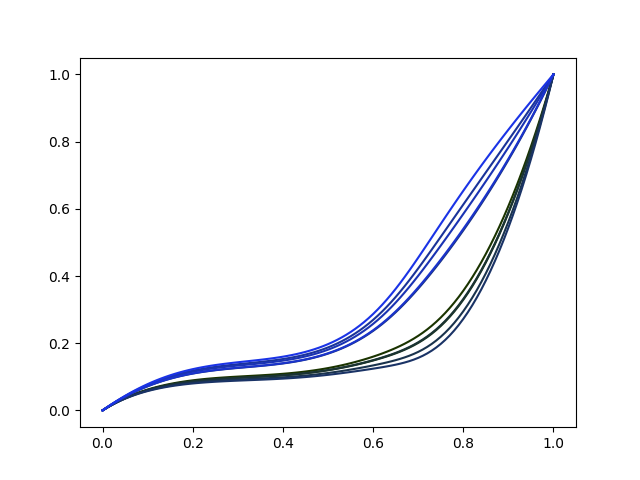}  \includegraphics[width=0.265\linewidth]{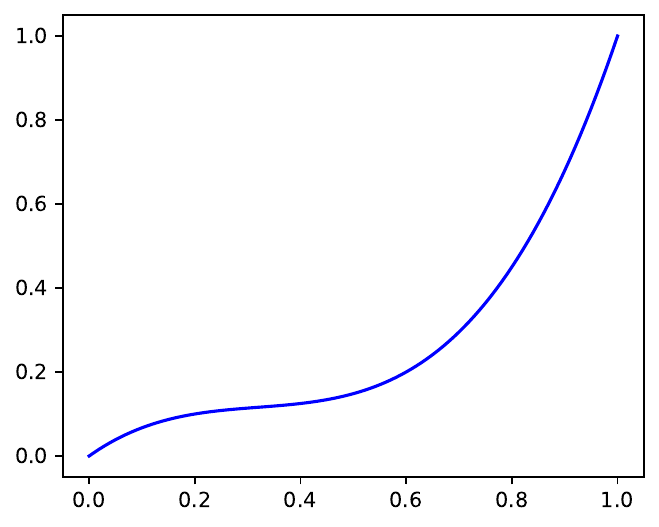} = 
\includegraphics[width=0.32\linewidth]{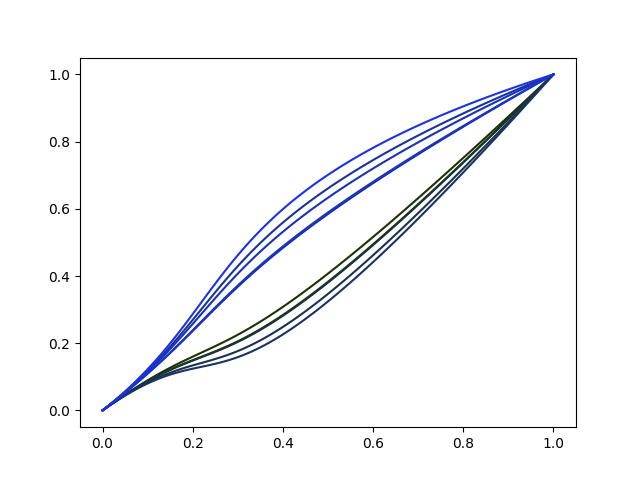}
    \caption{\textit{Centering method: $\bm F_{\widetilde{\bmu}}^{-1} = \bm F_{ \bmu }^{-1} \circ F_{\oplus}$.} On the left are realizations of random function $\bm F^{-1}_{\bm\mu}$, denoted by $F_t^{-1}, t =1, ..., 10$. On the middle is $F_{\oplus}^{-1}$. On the right are $F_t^{-1}\circ F_{\oplus}, t =1, ..., 10$, representing the realizations of $\bm F_{\widetilde{\bmu}}^{-1}$. We can see that the point-wise mean in the right subfigure becomes the identity function.}
    \label{fig: center}
\end{figure}

Let $\bm\mu \in \WR$ a random probability measure which is square integrable thus admits a Fréchet mean, $\mathbb{E}_\oplus\bm\mu$. According to Definition of Fréchet mean in \ref{def: frechet mean}, we have 
\begin{equation}
    F_{\oplus}^{-1}(p) = \mathbb{E}\left[\bm F_{\bmu}^{-1}(p) \right], \; p \in (0, 1),
\end{equation}
where $F_{\oplus}^{-1}$ and $\bm F_{\bmu}^{-1}$ are respectively quantile functions of $\mathbb{E}_\oplus\bm\mu$ and $\bmu$. Denote the quantile function of the transformed measure by $\bm F_{\widetilde{\bmu}}^{-1}$. The centering to $\mathcal{U}(0,1)$ requires that 
\begin{equation}
\mathbb{E}\left[\bm F_{\widetilde{\bmu}}^{-1}(p) \right] = p, \; p \in (0, 1).
\end{equation}
Note that the composition function $\bm F_{ \bmu }^{-1} \circ F_{\oplus},$ with $F_{\oplus}$ the cdf of $\mathbb{E}_\oplus\bm\mu$ satisfies 
\begin{equation}
\mathbb{E}\left[ \bm F_{ \bmu }^{-1} \circ F_{\oplus}(p) \right] = F_{\oplus}^{-1}\left( F_{\oplus}(p) \right) = p, \; p \in (0, 1),
\end{equation}
when $F_{\oplus}$ is strictly increasing. This motivates us to define $\bm F_{\widetilde{\bmu}}^{-1}$ as $\bm F_{ \bmu }^{-1} \circ F_{\oplus}$. However there are two technical issues. First, $\bm F_{ \bmu }^{-1} \circ F_{\oplus}$ is not necessarily a quantile function, since its domain is not $(0,1)$. Second, when $F_{\oplus}$ fails to be strictly increasing, $F_{\oplus}^{-1}\circ F_{\oplus} \neq i$. Recall that a quantile function is defined as the left continuous inverse of its cdf. Therefore, we assume that 
\textit{$\bm\mu$ is supported over $[0,1]$ with a strictly increasing cdf $\bm F_{ \bmu }$}. The strict monotonicity of each realization $\bm F_{ \bmu }$ implies that of their Fréchet mean, $F_{\oplus}$. Therefore the transformed measure $\widetilde{\bmu}$ is defined via its quantile function $\bm F_{\widetilde{\bmu}}^{-1} = \bm F_{ \bmu }^{-1} \circ F_{\oplus}$. Figure \ref{fig: center} illustrates the centering method. 
Applying this method on distributions in the time series, $\bm\mu_{t}^i$, leads to Assumption \ref{assump: common frechet mean}. 
\begin{assump}\label{assump: common frechet mean}
For any $i=1,...,N, t \in \mathbb{Z}$,  $\bm\mu_{t}^i$ is supported over $[0,1]$ with a strictly increasing cdf $\bm F_{i,t}$.
\end{assump}

Thus, the centered measures $\widetilde{\bm\mu}_{t}^i$ we use in Model \eqref{eq: multi_AR_Wass} are defined by 
\begin{equation}\label{eq: trans_TS}
  \widetilde{\bm F}_{i,t}^{-1} = \bm F_{i,t}^{-1} \circ F_{i,\oplus},  
\end{equation}
where $ \widetilde{\bm F}_{i,t}^{-1}, \bm F_{i,t}^{-1}$ are respectively the quantiles of $\widetilde{\bm\mu}_{t}^i, \bm\mu_{t}^i$, and $F_{i,\oplus}$ is the cdf of $\mu_{i,\oplus}$. 

\subsection{Identifiability}\label{sec: identifiability}
In this section, we discuss the assumptions that  guarantee the identifiability of Model \eqref{eq: multi_AR_Wass}, and detail the form of random noise map $\bm\Gamma_{\bm\epsilon_t^i}$ which needs to take into account the identifiability as well. 

The model is so far not identifiable because the exponential map is not injective. Therefore, two different coefficient values $A_{ij}$ can possibly lead to the same value of the regression formula $\Exp_{Leb}\left\{\sum_{j=1}^N A_{ij}(\widetilde{\bm T}^{j,t-1}_{Leb}-id)\right\}$. Note that when restricting $\Exp_{Leb}$ on its log image of the Wasserstein space $\Log_{Leb}\WR$, the former becomes injective. Therefore, we impose that $\sum_{j=1}^N A_{ij}(\widetilde{\bm T}^{j,t-1}_{Leb}-id) \in \Log_{Leb}\WR$. The implying constraint on $A_{ij}$ is not explicit. To explicit it, we recall the following property. 
\begin{prop}
Let $g \in \Tan_{\gamma}$, then $g \in \Log_{\gamma}\WR$ if and only if $g + id$ is nondecreasing, $\gamma$-almost everywhere. 
\end{prop}
Therefore, given that $\widetilde{\bm T}^{j,t-1}$ can be any nondecreasing functions, requiring $\sum_{j=1}^N A_{ij}(\widetilde{\bm T}^{j,t-1}_{Leb}-id) + id$ is nondecreasing results in Assumption \ref{assump: B1+}. 
\begin{assump}\label{assump: B1+}
$\sum_{j=1}^NA_{ij} \leq 1, \, \forall i = 1, ..., N,$ and $0 \leq A_{ij} \leq 1, \, \forall i,j = 1, ..., N$.
\end{assump}
Similar assumptions are imposed in related works, see e.g.\ \cite[Assuption (A1)]{chen2021wasserstein} and \cite[Assumption (A3)]{petersen2019wasserstein}, to keep regression formulas in a logarithmic image.

Now, we discuss $\bm\Gamma_{\bm\epsilon_t^i}$. Two ways can be thought of to add randomness to the regression formula $\Exp_{Leb}\left\{\sum_{j=1}^N A_{ij}(\widetilde{\bm T}^{j,t-1}_{Leb}-id)\right\}$ through a random function $\bm\epsilon_t^i$: in the tangent space or in the Wasserstein space, respectively meaning $$\Exp_{Leb}\left\{\sum_{j=1}^N A_{ij}(\widetilde{\bm T}^{j,t-1}_{Leb}-id) + \bm\epsilon_t^i\right\} \mbox{ and } \bm\epsilon_t^i\#\Exp_{Leb}\left\{\sum_{j=1}^N A_{ij}(\widetilde{\bm T}^{j,t-1}_{Leb}-id)\right\},$$
where $\#$ is pushforward given in Definition \ref{def: exp}. 
The tangent space way will make $\sum_{j=1}^N A_{ij}(\widetilde{\bm T}^{j,t-1}_{Leb}-id) + \bm\epsilon_t^i$ leave the log image, thus we need to reconsider the model constraint to guarantee the identifiability. While taking into account additionally $\bm\epsilon_t^i$, it is not sure that we can derive explicit constraint on $A_{ij}$ as Assumption \ref{assump: B1+}. By contrast, the pushforward way has been proposed several times in literature \citep{petersen2019frechet,chen2021wasserstein}, and it will not affect the identifiability of the model. Thus, we adopt the form of pushforward. We need furthermore requirements on the noise to make it ``white''. We present these requirements in the complete formulation of the proposed model. 
\begin{definition}\label{def: wmar}
Let $\bm\mu_{t}^i, i =1, ..., N, t \in \mathbb{Z}$ be a collection of time series of probability measures in $\WR$, which satisfies Assumptions \ref{assump: stationary N means} and \ref{assump: common frechet mean}. They are said to follow a Wasserstein multivariate AR model, if their centered measures $\widetilde{\bm\mu}_{t}^i$ follow the system given  Assumption \ref{assump: B1+}: 
\begin{equation}\label{eq:wmar}
    \widetilde{\bm\mu}_{t}^i = \bm\epsilon_t^i\#\Exp_{Leb}\left\{\sum_{j=1}^N A_{ij}(\widetilde{\bm F}_{{j,t-1}}^{-1}-id)\right\},
\end{equation}
where $\widetilde{\bm F}_{{j,t-1}}^{-1}$ is the quantile function of the centered observation $\widetilde{\bm\mu}_{t-1}^j$, $\{\bm \epsilon_{i,t}\}_{i,t}$ are i.i.d. random\footnote{We do not consider degenerate distributions.} noise functions taking values in the space of extended quantile functions 
$$
\begin{aligned}
&\Pi = \{F^{-1}: [0, 1] \rightarrow [0, 1], \mbox{ such that } F^{-1}\big|_{(0,1)} \in \Log_{Leb}\WR + id, \\
&F^{-1}(0) := \inf\{x \in [0, 1]: F(x) > 0\}, \mbox{ and } F^{-1}(1) := \sup\{x \in [0, 1]: F(x) < 1\}\},
\end{aligned}
$$
endowed with $\|\cdot\|_{Leb}$ and the induced Borel algebra,
$\bm \epsilon_{i,t}$ is almost surely independent of $\bm \mu_{t-1}^i, \, i = 1, \ldots, N, $ for all $t \in \mathbb{Z}$, 
and 
\begin{equation}\label{eq: noise mean}
\mathbb{E}\left[\bm \epsilon_{i,t}(p)\right] = p, \, p \in [0,1].    
\end{equation}
\end{definition}
Construction in Equation \eqref{eq: noise mean} is an extension of $\mathbb{E}\bm \epsilon_{i,t} = 0$ in the standard VAR models, given that the notion of nullity now is represented by the identity function equivalently $\mathcal{U}(0,1)$. Additional requirement $\bm \epsilon_{i,t}(p) \in \Pi$ is to allow an explicit calculation of $\mathbb{E}_\oplus \widetilde{\bm\mu}_{t}^i.$
\begin{equation}
\begin{aligned}
\widetilde{\bm\mu}_{t}^i = &\bm\epsilon_t^i\#\Exp_{Leb}\left\{\sum_{j=1}^N A_{ij}(\widetilde{\bm F}_{{j,t-1}}^{-1}-id)\right\} = \bm\epsilon_t^i\# \left\{\sum_{j=1}^N A_{ij}(\widetilde{\bm F}_{{j,t-1}}^{-1}-id) + i \right\} \# Leb \\
&= \bm\epsilon_t^i\circ \left\{\sum_{j=1}^N A_{ij}(\widetilde{\bm F}_{{j,t-1}}^{-1}-id) + i \right\} \# Leb    
\end{aligned}
\end{equation}
If $\bm\epsilon_t^i \in \Pi$ then the composition $\bm\epsilon_t^i\circ \left\{\sum_{j=1}^N A_{ij}(\widetilde{\bm F}_{{j,t-1}}^{-1}-id) + id \right\}$ is a valid quantile function under Assumption \ref{assump: B1+}, which defines the quantile of $\widetilde{\bm\mu}_{t}^i$. It follows that the quantile of $\mathbb{E}_\oplus \widetilde{\bm\mu}_{t}^i$ equals $\mathbb{E} \bm\epsilon_t^i\circ \left\{\sum_{j=1}^N A_{ij}(\widetilde{\bm F}_{{j,t-1}}^{-1}-id) + id \right\}$. When $\bm\epsilon_t^i \notin \Pi$, the calculation of $\mathbb{E}_\oplus \widetilde{\bm\mu}_{t}^i$ is not tractable. 

An example of the noise function satisfying Definition \ref{def: wmar} as well as Assumption \ref{assump: lipschitz noise in expectation} imposed later, can be found in \cite[Equation  (38)]{chen2021wasserstein}. %However, in these works, not many examples of valid random distortion functions which satisfy the conditions in Equation (\ref{eq:wmar}) are given. 
To furthermore demonstrate that the conditions imposed on the noise function are not restrictive, we describe, in Section \ref{sec: num_exp} on numerical experiments, a general mechanism to generate the random noise.

\subsection{Model representation in quantile functions}
Given Assumption \ref{assump: B1+}, Model \eqref{eq:wmar} admits a representation in terms of quantile functions: 
\begin{equation}\label{eq: wmar_F}
    \widetilde{\bm F}_{i,t}^{-1} = \bm \epsilon_{i,t} \circ \left[ \sum_{j=1}^NA_{ij}\left(\widetilde{\bm F}_{j,t-1}^{-1} - id\right) + id \right],
    %, \quad Leb-\mbox{almost everywhere}
    \quad t \in \mathbb{Z}, \, i = 1, \dots, N,
\end{equation}
where $\widetilde{\bm F}_{i,t}^{-1}$ is the quantile function of centered measure $\widetilde{\bm \mu}_t^i$.
When reducing to the univariate case ($N = 1$), Model \eqref{eq: wmar_F} is similar to the AR model proposed in \cite[Model (4)]{zhu2023autoregressive}, with a regression coefficient constrained to belong to $(0,1]$. Point-wise, the model representation \eqref{eq: wmar_F} can be seen as a special case of standard VAR model. Note that for any $p(0,1)$, we have 
\begin{equation}
\mathbb{E}\left[\widetilde{\bm F}_{i,t}^{-1}(p) \bigg | \widetilde{\bm F}_{j,t-1}^{-1} \right] =  p + \sum_{j=1}^NA_{ij}\left(\widetilde{\bm F}_{j,t-1}^{-1}(p) - p\right),
    %, \quad Leb-\mbox{almost everywhere}
    \quad t \in \mathbb{Z}, \, i = 1, \dots, N,
\end{equation}
with $p = \mathbb{E}\left[\widetilde{\bm F}_{i,t}^{-1}(p)\right].$ The constraints of quantile functions lead to the ones of $A_{ij}$. This representation helps us to understand the proposed Wasserstein model. It moreover serves as a main tool to study the second order stationarity of the original model in next section. 
\color{black}
\subsection{Existence, uniqueness and second order stationarity}\label{sec: Existence, uniqueness and stationarity}

In this section, we study the second order stationarity of the proposed model. A convenient direction is studying the second order stationarity of its quantile representation \eqref{eq: wmar_F}. Because it can be seen as an iterated random functions (IRF) system. There are general results on the stationarity of an IRF in literature \citep{wu2004limit} that we can apply. Thus we focus on the representation \eqref{eq: wmar_F} in this section. 

An IRF is defined in a metric space, we first introduce the following product space for representation \eqref{eq: wmar_F}: 
$$
\left(\mathcal{X}\,,\, d\right) := \left(\mathcal{T}\,,\, \|\cdot\|_{Leb}\right)^{\otimes N},
$$
where $\mathcal{T} := \Log_{Leb}\WR + id$ is the space of all quantile functions of $\WR$, equipped with the norm $\|\cdot\|_{Leb}$ in the  tangent space  at $\mathcal{U}(0,1)$. 
Thus, we have 
\begin{equation}\label{eq: def_d}
d(\mathbf{X},\mathbf{Y}) := \sqrt{\sum\limits_{i = 1}^N\left\|\mathbf{X}_i - \mathbf{Y}_i\right\|_{Leb}^2}, \quad \mathbf{X} = (\mathbf{X}_i)_{i = 1}^N \in \mathcal{X},\; \mathbf{Y} = (\mathbf{Y}_i)_{i = 1}^N \in \mathcal{X}.   
\end{equation}
Representation (\ref{eq: wmar_F}) can be interpreted as an IRF system operating on the state space $\left(\mathcal{X}\,,\, d\right)$, written as
\begin{equation}\label{eq: iter_sys}
    \X_{t} = \bm \Phi_{ \epsilon_t}\left(\X_{t-1}\right),
    %, \quad Leb-\mbox{almost everywhere}
\end{equation}
where $\mathbf{X}_{t} = (\mathbf{X}_{i,t})_{i = 1}^N, \; {\bm \epsilon}_{t} = ({\bm \epsilon}_{i,t})_{i = 1}^N$, and $\bm \Phi_{ \epsilon_t}\left(\X_{t-1}\right) = (\bm \Phi^i_{ \epsilon_t}\left(\X_{t-1}\right))_{i = 1}^N$ with
$$\bm \Phi^i_{ \epsilon_t}\left(\X_{t-1}\right) := \bm \epsilon_{i,t} \circ \left[ \sum_{j=1}^NA_{ij}\left(\X_{j,t-1} - id\right) + id \right].$$ We first study the existence and the uniqueness of the solution to the IRF system in the metric space $\left(\mathcal{X}\,,\, d\right)$.

For time series models in a Hilbert space, two standard assumptions that ensure the existence and the uniqueness of the solutions are the boundedness of the $L_p$ norm of random  additive noise and the contraction of the regression operator. For Model \eqref{eq: iter_sys}, the random noise $\bm\epsilon_{i,t}$ is bounded between $0$ and $1$, thus  $\mathbb{E}\left[d^p\left(\X, {\bm \epsilon} \right)\right]$ is bounded for all $\X \in \mathcal{X}$, which is the $L_p$ norm equivalent condition in the metric space setting. Then, to have a contractive map $\bm \Phi_{ \epsilon_t}$, we shall rely on an interplay between $A_{ij}$ and $\bm \epsilon_{i,t}$, since it is applied in a nonlinear way. More specifically, we impose Assumptions \ref{assump: lipschitz noise in expectation} and \ref{assump: l2 norm on TS coefficient} below on Model \eqref{eq: iter_sys}. 
\begin{assump}\label{assump: lipschitz noise in expectation} There exists a constant $L > 0$ such that
$\mathbb{E}\left[\bm \epsilon_{i,t}(x) - \bm \epsilon_{i,t}(y)\right]^2 \leq L^2(x-y)^2, \; \forall x, y \in [0, 1], \quad t \in \mathbb{Z}, \, i = 1, \dots, N$,
\end{assump}
\begin{assump}\label{assump: l2 norm on TS coefficient}
The matrix of AR coefficients satisfies $\|A\|_2 < \frac{1}{L}$, where $L$ is the Lipschitz constant from Assumption \ref{assump: lipschitz noise in expectation}.
\end{assump}
% In A3 A4 2 can be replaced by other p, but accordingly we should use Lp as well in Tangent space and in fact wasserstein distance, then the whole Thm 2.1 holds in Lp. Moreover, because Lp convergence -> Lq convergence when p > q, thus the lightest assumption is to consider L1. But this does not permits the underlying Hilbert space, which we require d is an induced Hilbert norm. This if we study furthermore stationarity, only L2 is allowed. 
Note that, Assumption \ref{assump: lipschitz noise in expectation} implies that $\bm \epsilon_{i,t}$ is $L$-Lipschitz in expectation. For increasing functions from $[0, 1]$ to $[0, 1]$, the smallest value of $L$ is 1 that is attained by the identity function. Therefore, Assumption  \ref{assump: l2 norm on TS coefficient} implies that $\|A\|_2 < 1$, which is the usual contraction assumption for standard VAR models in an Euclidean space. We now state the existence and uniqueness results.

\begin{theorem}\label{thm: existence_uniqueness} 
Under Assumptions \ref{assump: B1+}, \ref{assump: lipschitz noise in expectation} and \ref{assump: l2 norm on TS coefficient}, the IRF system \eqref{eq: iter_sys} almost surely admits a solution $\X_{t}, \; t \in \mathbb{Z}$, with the same marginal distribution $\pi$, namely, $\X_{t} \stackrel{d}{=} \pi, \;\forall \; t \in \mathbb{Z}$, where the notation $\stackrel{d}{=}$ means equality in distribution. Moreover, if there exists another solution ${\bm S}_{t}, \; t \in \mathbb{Z}$, 
%such that 
% $$
% \mathbb{E}\left[d^{\beta}({\bm S}_{t}, Z^{0})\right] < \infty,
% $$
% for some $\beta > 0$, and some $Z^{0} \in \mathcal{X}$  (thus for all), 
% S_t is bounded function because of (X,d)
then for all $t \in \mathbb{Z}$
$$
\X_{t} \stackrel{d}{=} {\bm S}_{t},   \;  \mbox{ almost  surely}. 
$$
\end{theorem}

Theorem \ref{thm: existence_uniqueness} states that under Assumptions \ref{assump: lipschitz noise in expectation} and \ref{assump: l2 norm on TS coefficient}, a well defined IRF system \eqref{eq: iter_sys} (namely when Assumption \ref{assump: B1+}  is satisfied) permits a unique solution in $\left(\mathcal{X}\,,\, d\right)$ almost surely. 
%Moreover, such solution is stationary in the sense that $\X_t \stackrel{d}{=} \pi, \, t \in \mathbb{Z}$. However, the stationarity in weak sense for time series requires additionally the auto-covariance to be time-invariant. 
Next, we show that this solution is furthermore stationary as a functional time series in a Hilbert space. 
To this end, we need to assume that there is an underlying Hilbert space associated to $\left(\mathcal{X}\,,\, d\right)$, with $\mathcal{X}$ its subset and $d$ equal to the induced norm of its inner product. Such Hilbert space exists
$$
(\Tan_{Leb}+id,\langle,\rangle_{Leb})^{\otimes N},
$$
\CB{which is furthermore separable.}
Thus we have $$
\langle X,Y \rangle = \sum\limits_{i = 1}^N\left\langle X_i, Y_i\right\rangle_{Leb}, \; X, Y \in (\Tan_{Leb}+id,\langle,\rangle_{Leb})^{\otimes N}.$$

%Thus, $\left(\mathcal{X}\,,\, \langle\cdot,\cdot\rangle\right)$ is the product Hilbert space of $N$ tangent space $\left(\mathcal{X}\,,\, \langle\cdot,\cdot\rangle_{Leb}\right)$.
We recall the conventional definition of second order stationarity for process in a separable Hilbert space, see for example \cite[Definition 2.2]{zhang2021wasserstein}. We recall the definition. 
%since we need to use the results before in d, thus we need to assume the inner product is compatible with d then by cauchy-schwarz, we can use the previous results.

\begin{definition}\label{def: stationarity hilbert space}
A random process $\{\mathbf{V}_t\}_t$ in a separable Hilbert space $\left(\mathcal{H}, \langle\cdot,\cdot\rangle\right)$ is said to be stationary if the following properties are satisfied.
\begin{enumerate}
    \item $\mathbb{E}\left\|\mathbf{V}_t\right\|^2 < \infty$.
    \item The Hilbert mean $U := \mathbb{E}\left[\mathbf{V}_t\right]$ does not depend on $t$.
    \item The auto-covariance operators defined as \begin{equation}
        \mathcal{G}_{t,t-h}(V) := \mathbb{E}\left\langle\mathbf{V}_{t}-U,V\right\rangle\left(\mathbf{V}_{t-h}-U\right), \quad V \in \mathcal{H}, 
    \end{equation}
    do not depend on $t$, that is $ \mathcal{G}_{t,t-h}(V) =  \mathcal{G}_{0,-h}(V)  $ for all $t$.
\end{enumerate}
\end{definition}
Then, Theorem \ref{thm: stationarity} below gives the second order stationarity result.
\begin{theorem}\label{thm: stationarity}
The unique solution given in Theorem \ref{thm: existence_uniqueness} is stationary as a random process in $\left(\mathcal{X}\,,\, \langle\cdot,\cdot\rangle\right)$ in the sense of Definition \ref{def: stationarity hilbert space}.
\end{theorem}

Finally, we point out in Proposition \ref{prop: representation A}
%, \ref{prop: Lipschitz solution} 
additional properties of  the IRF system \eqref{eq: iter_sys} that will serve in the following section of the estimation of the matrix of coefficients in  Model (\ref{eq: wmar_F}). More properties of the stationary solution see the supplemental materials. 
\begin{prop}\label{prop: representation A}
Given the stationary solution $\X_t$ of the IRF system \eqref{eq: iter_sys}, we define matrices $\Gamma(0), \Gamma(1) \in \R^{N \times N}$ as
\begin{equation}
    \begin{aligned}
    [\Gamma(0)]_{j,l} &=\mathbb{E}\, \langle\X_{j,t-1} - id \, , \, \X_{l,t-1} - id\rangle_{Leb} \\
    [\Gamma(1)]_{j,l} &=\mathbb{E}\, \langle\X_{j,t} - id \, , \, \X_{l,t-1} - id\rangle_{Leb},
    \end{aligned}
\end{equation}
for $1 \leq j, l \leq N$. We have 
\begin{enumerate}
    \item $\Gamma(0)$ is nonsingular,
    \item the coefficient matrix $A = [A_{ij}]$ of the IRF system \eqref{eq: iter_sys} admits the representation
\begin{equation}\label{eq: representation A}
A = {\Gamma}(1)\left[{\Gamma}(0)\right]^{-1}.
\end{equation}
\end{enumerate}
\end{prop}
Expression \eqref{eq: representation A} for the matrix $A$ is analogous to the one of VAR models, with matrices $\Gamma(0), \Gamma(1)$  carrying out the information on the correlation. However, compared to the auto-covariance operators in Definition \ref{def: stationarity hilbert space}, the matrices $\Gamma(0)$ and $\Gamma(1)$  rather reflect the average auto-covariance taking into account additionally the correlated level along the function domain.

% \begin{assump}\label{assump: lipschitz noise}
% $\left|\bm \epsilon_{i,t}(x) - \bm \epsilon_{i,t}(y)\right| \leq L|x-y|, \; \forall x, y \in [0, 1], \quad t \in \mathbb{Z}, \, i = 1, \dots, N$.
% \end{assump}
% \begin{prop}\label{prop: Lipschitz solution} 
% Given Assumption \ref{assump: lipschitz noise}, the unique solution $\X_{t}$ of System \eqref{eq: iter_sys} is $L$-Lipschitz continuous, for all $t \in \mathbb{Z}$, almost surely.
% \end{prop}
% Assumption \ref{assump: lipschitz noise} is a stronger assumption compared to Assumption \ref{assump: lipschitz noise in expectation}, which is not required by the existence, uniqueness and staionarity of the solution. However, it is needed to demonstrate the consistency of the proposed estimator of $A$ in the next section.

\section{Estimation of the regression coefficients}\label{sec: estimation}
In this section, we develop consistent estimators of coefficient $A$, given $T+1$ observed distributions ${\bm \mu}_t^i, \, t = 0, 1, \dots, T$ for each series $i = 1, \dots, N$. 
%consistency refers to convergence in proba
%by contrast, the asymptotic properties refer to consistency + clt 
We assume that the distributions are fully observed, instead of indirectly observed through their samples.

We propose to apply the least squares method on Model \eqref{eq:wmar}, corresponding to the following estimator: 
\begin{equation}\label{eq: exact lse}
\begin{aligned}
    \widetilde {\bm A}_{i:} = & \argmin_{A_{i:} } \frac{1}{T} \sum_{t = 1}^T d_{W}^2\left(\widetilde{\bm \mu}_t^i \, , \,  \Exp_{Leb}\left\{\sum_{j=1}^NA_{ij}(\tbF_{j,t-1}^{-1} - id)\right\}\right),
    \quad i = 1, \dots, N,  \\
    = &\argmin_{A_{i:} } \frac{1}{T} \sum_{t = 1}^T \left\| \, \tbF_{i,t}^{-1} -   \sum_{j=1}^NA_{ij}\left(\tbF_{j,t-1}^{-1} - id\right) - id \, \right\|_{Leb}^2,
    \quad i = 1, \dots, N. 
\end{aligned}
\end{equation}
The estimator is defined as 
the minimizer of the sum of squared residuals as usual. The difference here is the residuals are measured by the Wasserstein distance, instead of $l_2$ norm. 

In practice, we only observe $\bm \mu_t^i$. Thus we need to first center $\bm \mu_t^i$, then use the centered distributions $\widetilde{\bm \mu}_t^i$ to calculate the estimator. A detail is that the centering method proposed in Section \ref{sec: centering} uses the quantile function $F_{i,\oplus}^{-1}$ of Fréchet mean $\mathbb{E}_\oplus(\bm \mu_t^i)$, which is a population value not available in practice. Thus we need to estimate $F_{i,\oplus}^{-1}$, then use the estimation in the centering method. A natural estimator is the quantile function of empirical Fréchet mean. We recall the definition \citep{panaretos2020invitation} as below. 
\begin{definition}\label{def: empirical frechet}
Let $\mu_1, \dots, {\mu}_T$ be probability measures in $\WR$. The empirical Fr\'echet mean of $\mu_1, \dots, {\mu}_T$, denoted by $\Bar{\mu}$, is defined as the unique minimizer of 
\begin{equation}
    \Bar{\mu} = \min_{\nu \in \WR} \frac{1}{T} \sum_{t = 1}^Td^2_W({\mu}_t, \nu).
\end{equation}
\end{definition}
Analogously, we have  
\begin{equation}
    F_{\Bar{\mu}}^{-1}(p) = \frac{1}{T} \sum_{t = 1}^TF_{\mu_t}^{-1}(p), \; p \in (0, 1).
\end{equation}
Therefore, we first estimate  $F_{i,\oplus}^{-1}$ by 
\begin{equation}\label{eq: empirical frechet mean}
{\bm F}_{\Bar{\mu}_i}^{-1} =  \frac{1}{T} \sum_{t = 1}^T\bm F_{{i,t}}^{-1},   
\end{equation}
then center  ${\bm \mu}_{i,t}$ by ${\bm F}_{\Bar{\mu}_i}^{-1}$ as 
\begin{equation}\label{eq: trans_TS app}
  \widehat{\bm F}_{i,t}^{-1}  = {\bm F}_{i,t}^{-1}\circ {\bm F}_{\Bar{\mu}_i}.
\end{equation}
Quantile function $\widehat{\bm F}_{i,t}^{-1}$ defines the centered measure $\widehat{\bm \mu}_{i,t}$. We use a different notation to distinguish from $\widetilde{\bm \mu}_{i,t}$, the centered measure using population Fréchet mean.

Plug $\widehat{\bm \mu}_{i,t}$ into the least squares formula \eqref{eq: exact lse}, we obtain an approximate least squares estimator $\widehat{\bm A}_{o}$ whose rows satisfy
\begin{equation}\label{eq: app lse}
    \left[\widehat{\bm A}_{o}\right]_{i:} = \argmin_{A_{i:} } \frac{1}{T} \sum_{t = 1}^T \left\|\widehat{\bm F}_{i,t}^{-1} -   \sum_{j=1}^NA_{ij}\left(\widehat{\bm F}_{j,t-1}^{-1} - id\right) - id\right\|_{Leb}^2, \quad i = 1, \dots, N, 
\end{equation}
%Note that, since the empirical Fr\'echet means of all $\widehat{\bm \mu}_{i,t}$ are also uniform, similarly to the population setting, we do not need to consider the additional terms in the estimation problem to cancel out the unequal empirical Fr\'echet means.

Finally, we add the coefficient constraints \ref{assump: B1+} to the problem: 
\begin{equation}\label{eq: op L2}
    \widehat {\bm A}_{i:} = \argmin_{A_{i:} \in B_{+}^1} \frac{1}{T} \sum_{t = 1}^T \left\|\widehat{\bm F}_{i,t}^{-1} -   \sum_{j=1}^NA_{ij}\left(\widehat{\bm F}_{j,t-1}^{-1} - id\right) - id\right\|_{Leb}^2, \quad i = 1, \dots, N, 
\end{equation}
where $B_{+}^1$ is $N$-dimensional simplex, that is the nonnegative
orthant of the $\ell_1$ unit ball $B^1$ in $\R^N$. 
%An important advantage of this constraint is to promote sparsity in $\widehat {\bm A}_{i:}$, which will be illustrated in Section \ref{sec: num_exp}. 
To solve the optimization problem, we first notice that it admits a vector form: 
\begin{equation}\label{eq: vec_rep_cols}
    \widehat {\bm A}_{i:} = \argmin_{A_{i:} \in B_{+}^1} \frac{1}{2} {\bm A}_{i:} \left[\widehat{\bm \Gamma}(0)\right]{\bm A}_{i:}^\top - \left[\widehat{\bm \Gamma}(1)\right]_{i,:}{\bm A}_{i:}^\top, \quad i = 1. \dots, N,
\end{equation}
where $$[\widehat{\bm \Gamma}(0)]_{j,l} =\frac{1}{T}\sum_{t = 1}^T\langle\widehat{\bm F}_{j,t-1}^{-1} - id \, , \, \widehat{\bm F}_{l,t-1}^{-1} - id\rangle_{Leb}$$ and $$[\widehat{\bm \Gamma}(1)]_{j,l} =\frac{1}{T}\sum_{t = 1}^T\langle\widehat{\bm F}_{j,t}^{-1} - id \, , \, \widehat{\bm F}_{l,t-1}^{-1} - id\rangle_{Leb}.$$
Then problem \eqref{eq: vec_rep_cols} can be solved easily by standard frameworks such as accelerated projected gradient descent \cite[Chapter 4.3]{parikh2014proximal}. The projection onto $B_{+}^1$ also exists in literature, for example in \cite{thai2015projected}.

The consistency of the proposed estimator $\widehat {\bm A}$ is provided in Theorem \ref{thm: consistency}. The details of its proof is given in the supplemental material. 
\begin{theorem}\label{thm: consistency}
%(Consistency of the proposed estimator)
Assume that ${\bm \mu}_t^i, \, i = 1, \dots, N$ satisfy Assumption \ref{assump: common frechet mean} for $t = 0, 1, \dots, T$, and the transformed sequence $\widetilde{\bm F}^{-1}_{t}, \, t = 0, 1, \dots, T$ satisfies Model \eqref{eq: wmar_F} with Assumption \ref{assump: B1+} true. Suppose additionally that $\widetilde{\bm F}^{-1}_0 \stackrel{d}{=} \pi$ with $\pi$ the stationary distribution defined in Theorem \ref{thm: existence_uniqueness}. Given Assumptions \ref{assump: lipschitz noise in expectation} and \ref{assump: l2 norm on TS coefficient} hold true, and 
the true coefficient $A$ satisfies Assumption \ref{assump: B1+}, namely, $A_{i:} \in B_{+}^1, \,i = 1, \dots, N$, we have 
\begin{equation}
    \widehat{\bm A} - A \stackrel{p}{\rightarrow} 0.
\end{equation}
\end{theorem}
Note that, the result is provided without convergence rate, which requires stronger regularity on the random noise functions $\bm \epsilon_{i,t}$ than currently supposed in Assumption \ref{assump: lipschitz noise in expectation}, such as assuming $\mathbb{E}\{\sup_{x,y}\left[\bm \epsilon_{i,t}(x) - \bm \epsilon_{i,t}(y)\right]^2\} \leq L^2(x-y)^2, \; t \in \mathbb{Z}, \, i = 1, \dots, N$. This will make sure the same regularity holds for the stationary solutions $\tbF_{i,t}$. 
A stronger assumption can be $\left|\bm \epsilon_{i,t}(x) - \bm \epsilon_{i,t}(y)\right| \leq L|x-y|, \; \forall x, y \in [0, 1], \quad t \in \mathbb{Z}, \, i = 1, \dots, N, $ almost surely, implying that $\tbF_{i,t}$ is lipschitz almost surely. However such assumptions can be less realistic. Because the asymptotic results do not help the real applications, we do not consider additional assumptions to welcome a large variety of applicable data sets.

\section{Numerical experiments}\label{sec: num_exp}
In Section \ref{sec: simulations}, we firstly demonstrate the consistency result of the proposed estimator using synthetic data. Then, in Section \ref{sec: real data set population} we fit the model on the two real data sets: one of the age distributions for countries in the European Union, and the other of free slot distributions for public bike stations at Paris. The estimated coefficient matrix $\widehat{\bm A}$ allows us to understand the cross-country or cross-station dependency structures in their distributions. In particular, we visualize these learned structures using directed weighted graphs induced by $\widehat{\bm A}$.
% This visualization clearly exhibits the links among the sensors studied, and this supports our intention during the model designing to adopt a coefficient matrix instead of coefficient function or operator. 

\subsection{Simulations}\label{sec: simulations}

\subsubsection{Generation of the synthetic data}\label{sec: Procedures of generating the synthetic data}
We firstly propose a mechanism to generate random noises that satisfy Model \ref{def: wmar} and Assumption \ref{assump: lipschitz noise in expectation}. We consider the random functions defined by
\begin{equation}\label{eq: epsilon}
    \bm \epsilon_g = \frac{1+\bm\xi}{2}g\circ h^{-1} + \frac{1-\bm\xi}{2}h^{-1},
\end{equation}
where $g$ is a non-decreasing right-continuous function from $[0,1]$ to $[0,1]$, $h^{-1}$ is the left continuous inverse of $h = \frac{1}{2}(g + id)$ , and $\bm \xi \sim U(-1,1)$ is a random variable. For any given function $g$, we can sample a family of noise functions $\bm \epsilon_{i,t} \stackrel{i.i.d.}{\sim} \bm \epsilon_g$, when sampling $\bm \xi_{i,t} \stackrel{i.i.d.}{\sim} U(-1,1)$. This construction  of random noise functions is inspired by the one proposed in \cite[Equation (13)]{zhu2023autoregressive}. However, we have modified their construction of $h$ and of the random coefficients. 
%\CR{(not sure to put) Since we found in our experiments that the expectation of the random distortion function proposed in \cite[Equation (13)]{zhu2023autoregressive} is not the identity function as it should be. Thus we propose Formula \eqref{eq: epsilon} as a way of correction, which can  satisfy additionally our assumptions of $\bm \epsilon_g$ under the extra conditions on $g$.} 
It is easy to verify that 
\begin{equation}
 \mathbb{E} [\bm \epsilon_g] = \frac{1}{2}(g\circ h^{-1} + h^{-1}) = \frac{1}{2}(g + id)\circ h^{-1} = id.
\end{equation}
%\CR{Comment not put: I give the additional assumptions on $g$ here but not in the definition above to clarify these are not proposed in \cite[Equation (13)]{zhu2023autoregressive}.}

\begin{figure}[htbp]
    \centering
    \includegraphics[width=0.4\linewidth]{images/g.pdf}    \includegraphics[width=0.4\linewidth]{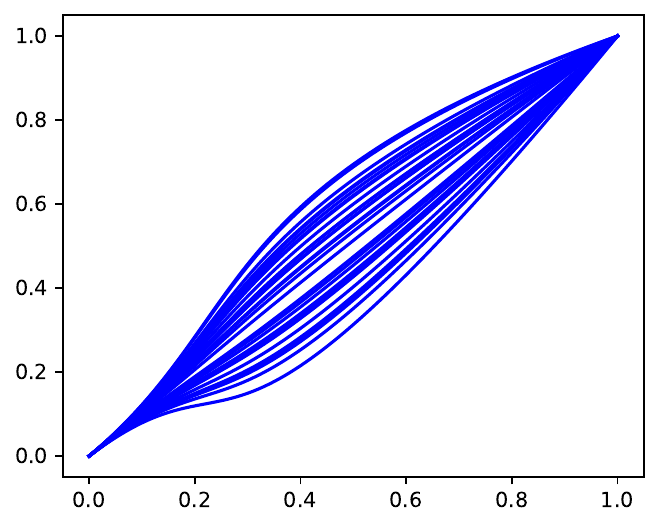}
    \caption{The function $g$ on the left is given by the natural cubic spline passing through the points $(0,0), (0.2,0.1), (0.6,0.2), (1,1)$. On the right is one realization of $30$ i.i.d. samples of the resulting $\bm \epsilon_g$.}
    \label{fig: g-epsilon}
\end{figure}

To make $\bm \epsilon_{i,t}$ satisfy additionally Model \eqref{eq:wmar} and Assumption \ref{assump: lipschitz noise in expectation}, we require $g$ to be furthermore continuous and differentiable. Then on the one hand, since $g$ is continuous and non-decreasing, any generated $\bm \epsilon_g$ is non-decreasing and left-continuous. On the other hand, note that 
\begin{equation}
\begin{aligned}
    [h^{-1}]' = \frac{1}{h'\circ h^{-1}} = \frac{1}{\frac{1}{2}(g'+ 1)\circ h^{-1}} = \frac{2}{g'\circ h^{-1}+ 1}.
\end{aligned}
\end{equation}
Thus, we have 
\begin{equation}
\begin{aligned}
    {\bm \epsilon_g}'&= \frac{1+\bm\xi}{2}\left(g'\circ h^{-1}\right) \frac{2}{g'\circ h^{-1}+ 1} + \frac{1-\bm\xi}{2}\frac{2}{g'\circ h^{-1}+ 1} \\
    & = \left(\frac{1+\bm\xi}{2}g'\circ h^{-1} + \frac{1-\bm\xi}{2}\right)\frac{2}{g'\circ h^{-1}+ 1} \\
    & = 1+\bm\xi - \bm\xi\frac{2}{g'\circ h^{-1}+ 1} = 1 + \bm\xi \left(1-\frac{2}{g'\circ h^{-1}+ 1}\right). 
\end{aligned}
\end{equation}
This implies 
$$
|{\bm \epsilon_g}'| \leq 1 + |\bm\xi| \left|1-\frac{2}{g'\circ h^{-1}+ 1}\right| \leq 2. 
$$
The bound comes from $\bm\xi \sim U(-1,1)$ and $g' \geq 0$, which is hence tight. Thus any $\bm \epsilon_g$ generated by Formula \eqref{eq: epsilon} is Lipschitz continuous, with the constant uniformly bounded by $2$ over $\bm \xi$. Note that Assumption \ref{assump: lipschitz noise in expectation} requires the Lipschitz continuity only in expectation. Thus, the i.i.d. samples $\bm \epsilon_{i,t}$ of any $\bm \epsilon_g$ satisfy obviously  Assumption \ref{assump: lipschitz noise in expectation} with the largest $L = 2$. Figure \ref{fig: g-epsilon} shows the function $g$ used in the simulation and one realization of $30$ i.i.d. samples of the resulting $\bm \epsilon_g$.

Secondly, we present the procedure to generate the true coefficient matrix $A$. We first generate a sparse matrix with the weights all positive in a random way, denoted by $A^0$. We then normalize each row of $A^0$ by the row sum to fulfill Assumptions \ref{assump: B1+}. We denote this last matrix still by $A^0$. Based on the previous mechanism for the random noise function, we take $L = 2$ in Assumption \ref{assump: l2 norm on TS coefficient}. Lastly, we scale down $A^0$ by $(2+\alpha)\|A^0\|_2$ to obtain a valid $A$. We test two values $\alpha = 0.1$ and $ \alpha =  0.5$ in our experiments.

Given a valid matrix of coefficients $A$ and samples of $\bm\epsilon_{i,t}$, we can then generate the ``centered'' quantile functions $\tbF^{-1}_{i,t}$ from Model \eqref{eq: wmar_F}. Note that, $\tbF^{-1}_{i,t}$ are only the simulations of the transformed data. Thus, we have to generate furthermore the population Fr\'echet mean $F^{-1}_{i,\oplus}$ of each univariate series in order to finally obtain the synthesized ``raw'' data, as the inverse of transformation \eqref{eq: trans_TS}:
$$
\bm F^{-1}_{i,t} = \tbF^{-1}_{i,t} \circ F_{i,\oplus}^{-1}.
$$
We set $F^{-1}_{i,\oplus}$ as the natural cubic spline of the points: $(0,0), (0.2,0.1), (0.6,0.2 + 0.2i/N), (1,1), \; i = 1, ..., N$.
The empirical Fr\'echet mean ${\bm F}^{-1}_{\Bar{\mu}_i}$ and the proposed estimator $\widehat{\bm A}$ are calculated on the synthesized ``raw'' data $\bm F^{-1}_{i,t}$. In Section \ref{sec: simulation results}, we will demonstrate the consistency result given in Theorem \ref{thm: consistency} with the synthetic data.

%$\tbF_0$ is set as $id$

%even demanded in the theorem assumption, we do not require $\tbF_0$ to follow the stationary distribution $\pi$, because when there is not analytic form for $\pi$, it is unrealistic in practice since we need to iterate infinite times of the IRF system. Also, indeed, this technical requirement exists also in VAR model, however, in practice, we do not try to fulfill it as other assumptions. 

\subsubsection{Experiment settings and  results}\label{sec: simulation results}
In this experiment, we demonstrate the consistency of the proposed estimator $\widehat{\bm A}$ for two different values $N =10$ and $N = 100$. For each $N$, we generate two true matrices $A$ for $\alpha = 0.1$ and $0.5$ respectively,  according to the procedure presented in Section \ref{sec: Procedures of generating the synthetic data}. With each $A$, we calculate the root mean square deviation (RMSD) successively  
\begin{equation}\label{eq: rmsd}
    \frac{\|\widehat{\bm A} - A\|_F}{\|A\|_F},
\end{equation}
with the synthetic data that it generates along time. To furthermore study the mean and the variance of the RMSD \eqref{eq: rmsd}, we run $100$ independent simulations for the same $A$. 

Note that the value of $\widehat{\bm A}$ we use in Equation \eqref{eq: rmsd} is the approximation obtained by the projected gradient descent applied to Problem \eqref{eq: op L2}. Thus the corresponding approximation error also accounts for the deviation which is on the order of the threshold we set in the stopping criteria. For all values of $N$, we use the same error threshold. We stop the algorithm as soon as the difference between the previous and the current updates in $\ell_2$ norm reaches $0.0001$, for the resolution of each row $\widehat{\bm A}_{i:}$. 

\begin{figure}[htbp]
    \centering
    \includegraphics[width=0.45\linewidth]{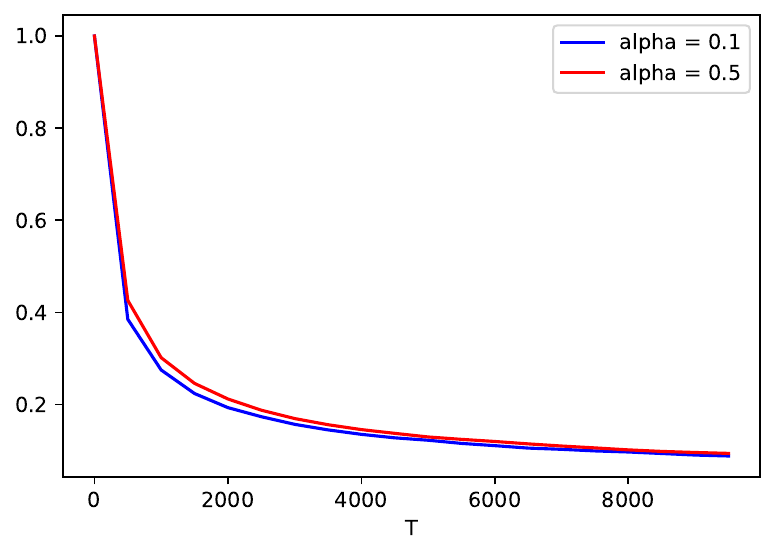}
    \includegraphics[width=0.47\linewidth]{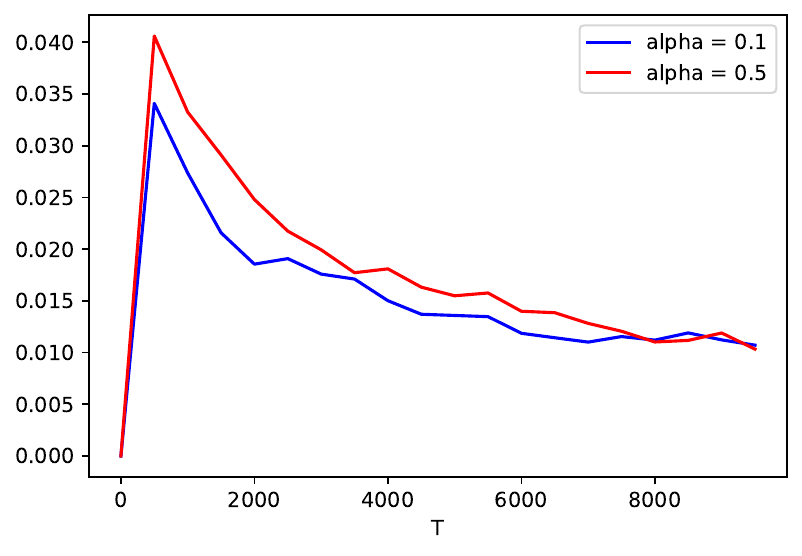}
    \caption{\textit{Mean (left) and standard deviation (right) of RMSD for $N = 10$.} The mean and the variance are calculated over $100$ simulations along time $T$, every $500$ time instants.}
    \label{fig: rmsd_N10}
\end{figure}

\begin{figure}[htbp]
    \centering
    \includegraphics[width=0.44\linewidth]{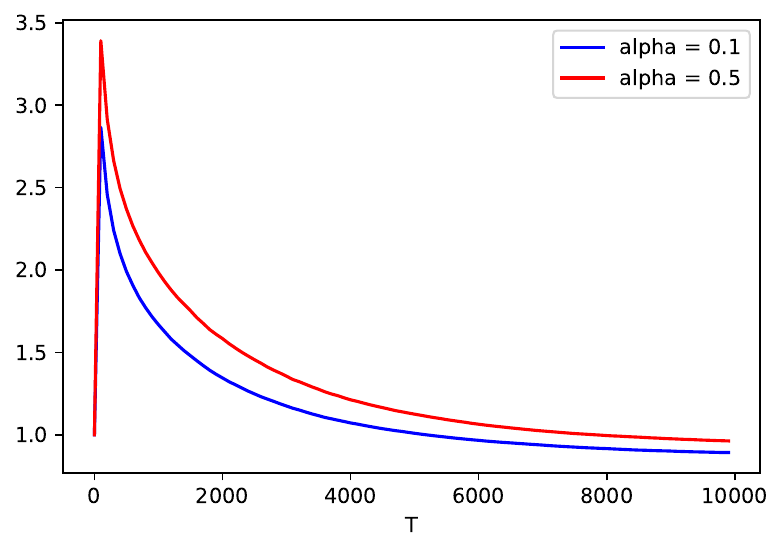}
    \includegraphics[width=0.46\linewidth]{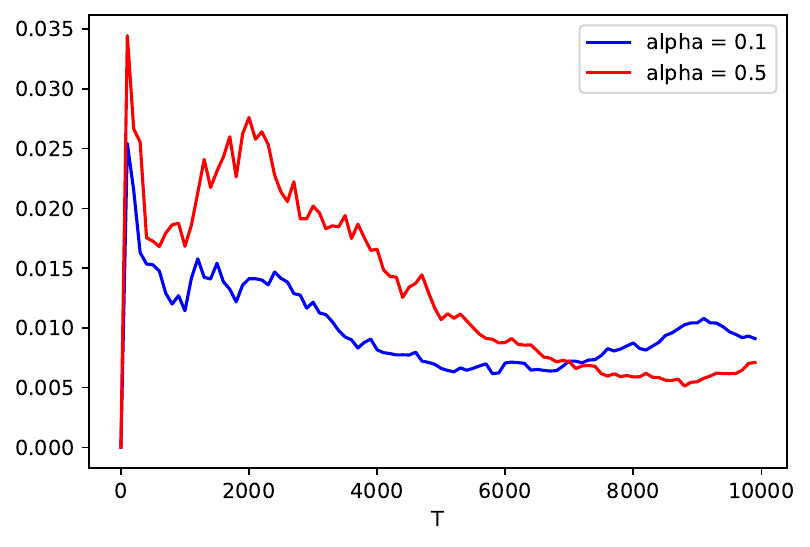}
    \caption{\textit{Mean (left) and standard deviation (right) of RMSD for $N = 100$.}  The mean and the variance are calculated every $100$ time instants.}
    \label{fig: rmsd_N100}
\end{figure}

We firstly show the evolution of RMSD for $N = 10, 100$ in Figures \ref{fig: rmsd_N10} and \ref{fig: rmsd_N100}, respectively. We can see that, all means and variances of the RMSD decrease towards zero as the sample size $T$ increases, for each $N$ and $\alpha$ value. \CB{This demonstrates empirically that $\widehat{\bm A}$ converges to $A$ in $L_2$, which implies hence the convergence in probability. Note that it is normal that the mean error is $10$ times greater for $N=100$ at $T=10000$ compared to $N=10$ because the increase in the number of $A_{ij}$ is quadratic with respect to the one of $N$. Thus $T=10000$ for $N=100$ is not yet in low dimension, while it is for $N=10$.}
%This demonstrates empirically that, when the model assumptions \ref{assump: B1+}, \ref{assump: lipschitz noise in expectation} and \ref{assump: l2 norm on TS coefficient} hold true for the data, the proposed estimator $\widehat{\bm A}$ converges to $A$ in probability, which is implied by the convergence of $\widehat{\bm A}$ to $A$ in $l_2$.

Additionally, we can notice that, the RMSD for $\alpha = 0.1$ which corresponds to larger $\ell_2$ norm of $A$ has a smaller mean in both cases, and also a smaller variance for most of the sample sizes $T$ investigated. 

\begin{figure}[htbp]
    \centering
    \includegraphics[width=0.45\linewidth]{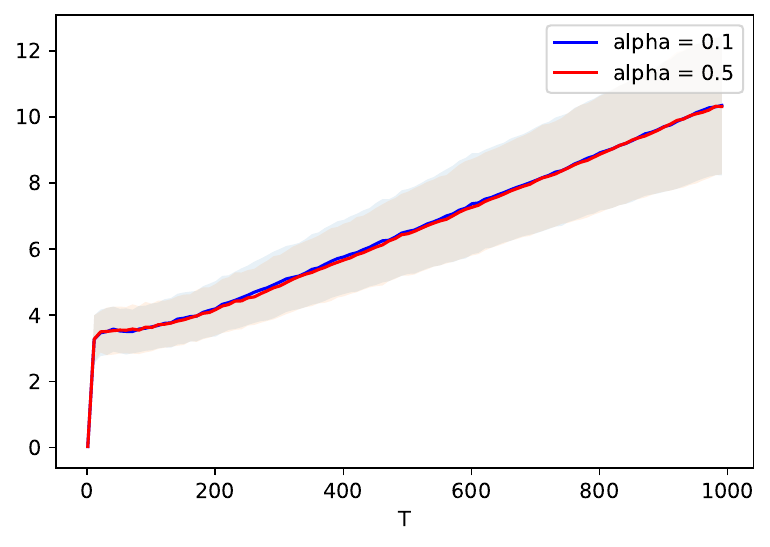}
    \includegraphics[width=0.47\linewidth]{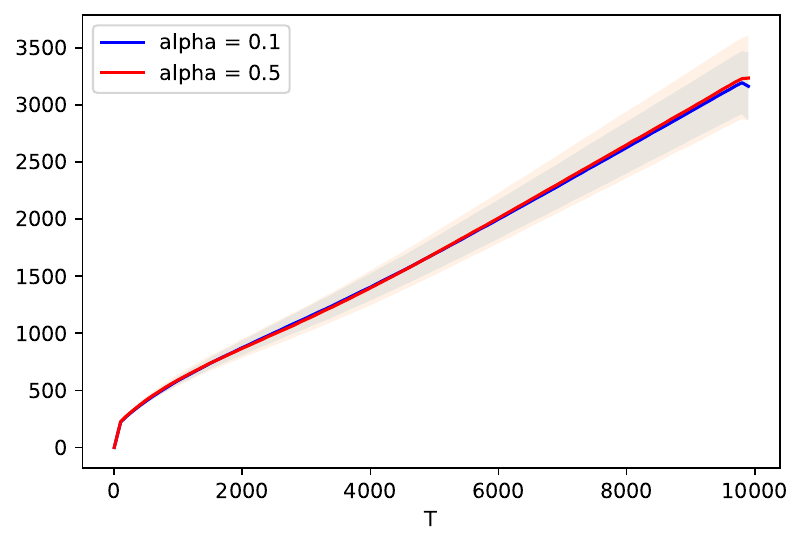}
    \caption{\textit{Calculation time (in seconds) of $\widehat{\bm A}$ with respect to the sample size $T$ for $N = 10$ (left) and $N = 100$ (right).} The calculation time counting starts from the computation of the empirical Fr\'echet means for Data transformation \eqref{eq: trans_TS app}, and ends when the accelerated projected gradient descent of  Problem \eqref{eq: op L2} finishes for the last row $i=N$.}
    \label{fig: rtime_N10}
\end{figure}

Lastly, we show in Figure \ref{fig: rtime_N10} the complete execution time of the model fitting on the raw data with respect to the sample size $T$. We can see that the execution time increases linearly with respect to $T$, and $A$ with the smaller $\ell_2$ norm requires sightly less time ($\alpha = 0.5$) than the other. The linear increase comes mainly from the loop over time $t = 1, ..., T$ in calculating the empirical Fr\'echet mean \eqref{eq: empirical frechet mean} and in calculating the matrices $\widehat{\bm \Gamma}(0), \widehat{\bm \Gamma}(1)$ in Equation \eqref{eq: vec_rep_cols}. The running time of these calculations is determined by the granularity in the numerical methods to approximate the  function composition, function inverse, and the inner product. The granularity applied during this simulation is $0.01$, that is we input/output only the quantile function values at grid $0, 0.01, ..., 0.99$ to/from each numerical approximation. 

%Impact of $N$: Note that the increase in the running time comes not only from the growth of dimension $N$, but also from the fact that we use the same error threshold to stop the gradient descent. Since the difference between updates is caused by more variables in larger dimension. 

\subsection{Age distribution of countries}\label{sec: real data set population}
We  test the proposed model with the real data set illustrated in Figure \ref{fig-intro: distributional} in the introduction. These data are from the US Census Bureau's International Data Base\footnote{The data that support the findings of this study are openly available in International Database (Time Series: various year - 2100) at \url{https://www.census.gov/data/developers/data-sets/international-database.html}.}, which provides the population estimates and projections for countries and areas by single year of age, over years. We would like to apply the proposed model on this international age distribution data to learn about the links among the age structures of different countries. Specially, we consider the countries and the micro-states in the European Union and/or Schengen Area. Because the corresponding data used during the model fitting starts in the $1990$s, we also include the former European Union member United Kingdom. Note that, Vatican City is not included since it is not available in the data base. Therefore, the list of $34$ countries in this study is: Austria, Belgium, Bulgaria, Croatia, Cyprus, Czech Republic, Denmark, Estonia, Finland, France, Germany, Greece, Hungary, Iceland, Ireland, Italy, Latvia, Liechtenstein,  Lithuania, Luxembourg, Malta, Monaco, Netherlands, Norway, Poland, Portugal, Romania, San Marino, Slovakia, Slovenia, Spain, Sweden, Switzerland, United Kingdom. Time-wise, we consider the $40$ years between $1996$ to $2035$. $1996$ is the earliest year for which the data for all the considered countries is available.

\begin{figure}[htbp]
    \centering

    \begin{subfigure}[t]{0.5\textwidth}
        \centering
        \includegraphics[width=\linewidth]{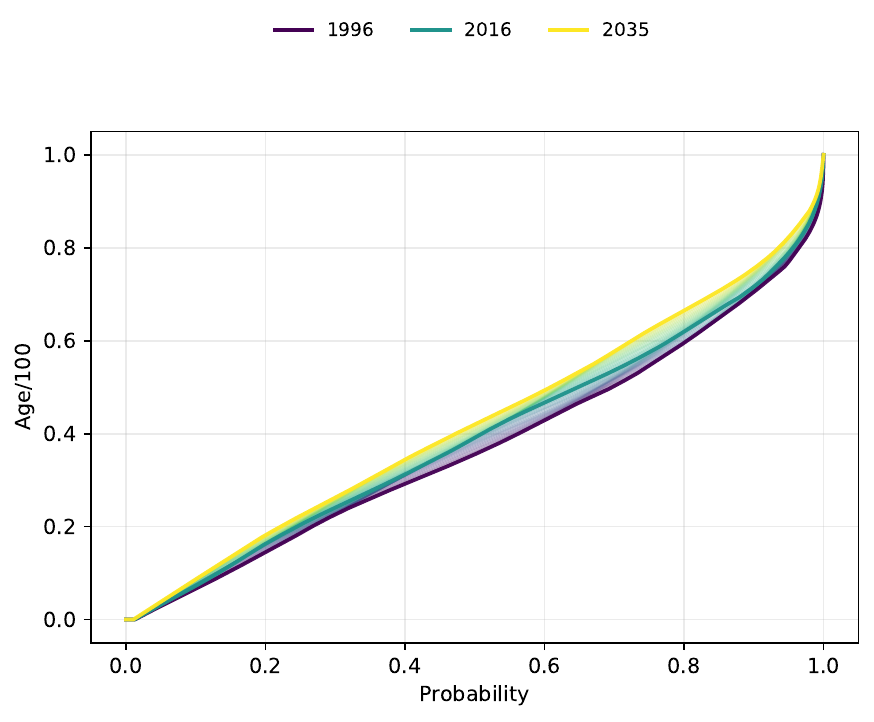}
        \caption{Raw series of quantile functions $\bm F_{i,t}^{-1}$}
        \label{fig:france_qt}
    \end{subfigure}
    \hfill
    \begin{subfigure}[t]{0.45\textwidth}
        \centering
        \includegraphics[width=\linewidth]{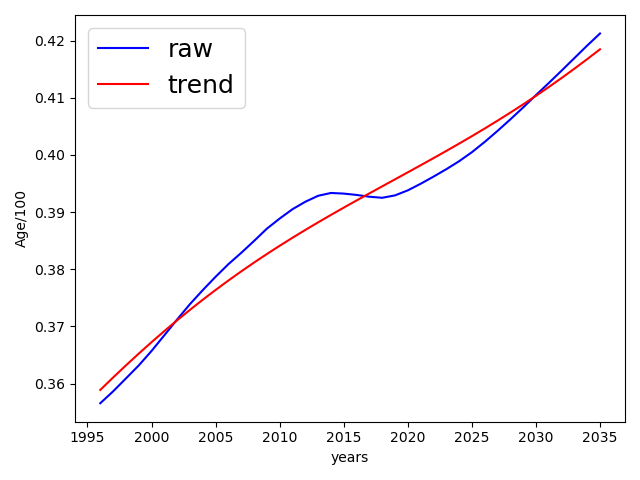}
        \caption{Raw series of $\bm F_{i,t}^{-1}(0.5)$ and its estimated trend $F_{i,\oplus}(t)^{-1}(0.5)$}
        \label{fig:france_qt_05}
    \end{subfigure}

    \vspace{0.5em}

    \begin{subfigure}[t]{0.48\textwidth}
        \centering
        \includegraphics[width=\linewidth]{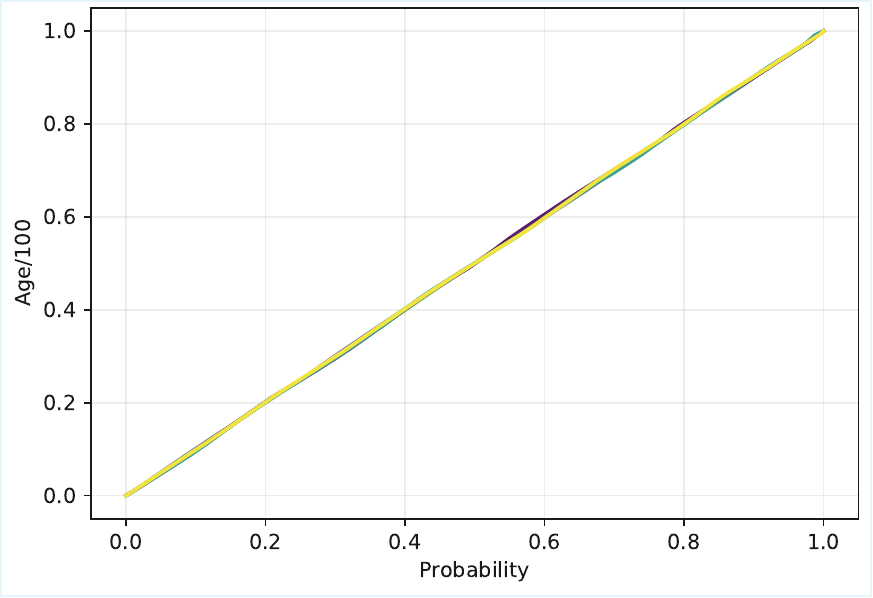}
        \caption{Detrended series of quantile functions $\bm F_{i,t}^{-1} \circ F_{i,\oplus}(t)$}
        \label{fig:france_qt_detr}
    \end{subfigure}
    \hfill
    \begin{subfigure}[t]{0.48\textwidth}
        \centering
        \includegraphics[width=\linewidth]{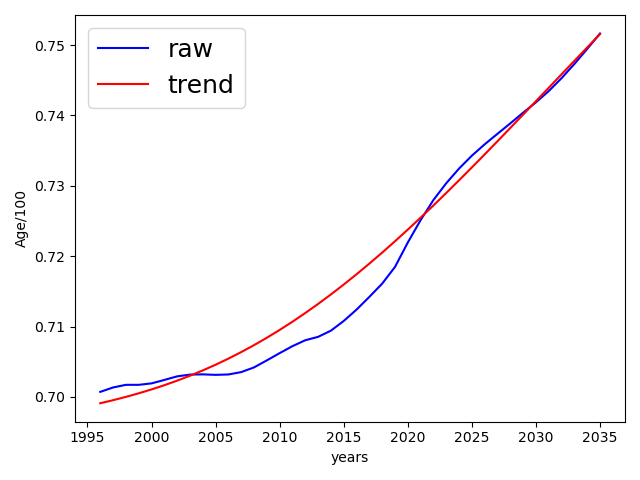}
        \caption{Raw series of $\bm F_{i,t}^{-1}(0.9)$ and its estimated trend $F_{i,\oplus}(t)^{-1}(0.9)$}
        \label{fig:france_qt_09}
    \end{subfigure}

    \caption{Detrending of the component series for France. In subfigure \ref{fig:france_qt},  one observes that the raw time series of quantile functions does not fluctuate around a fixed mean function. Instead, it gradually shifts toward the most recent quantile function, which corresponds to a more aged population. This trend is easier to see when the functions are projected onto single quantile levels (subfigures \ref{fig:france_qt_05} and \ref{fig:france_qt_09}). After detrending, the quantile functions oscillate around the identity function (subfigure \ref{fig:france_qt_detr}). The corresponding projected detrended series are shown in subfigures \ref{fig:france_qt_05} and \ref{fig:france_qt_09}.}
    \label{fig:detrending}
\end{figure}

\begin{figure}[htbp]
    \centering
    \includegraphics[width=0.7\linewidth]{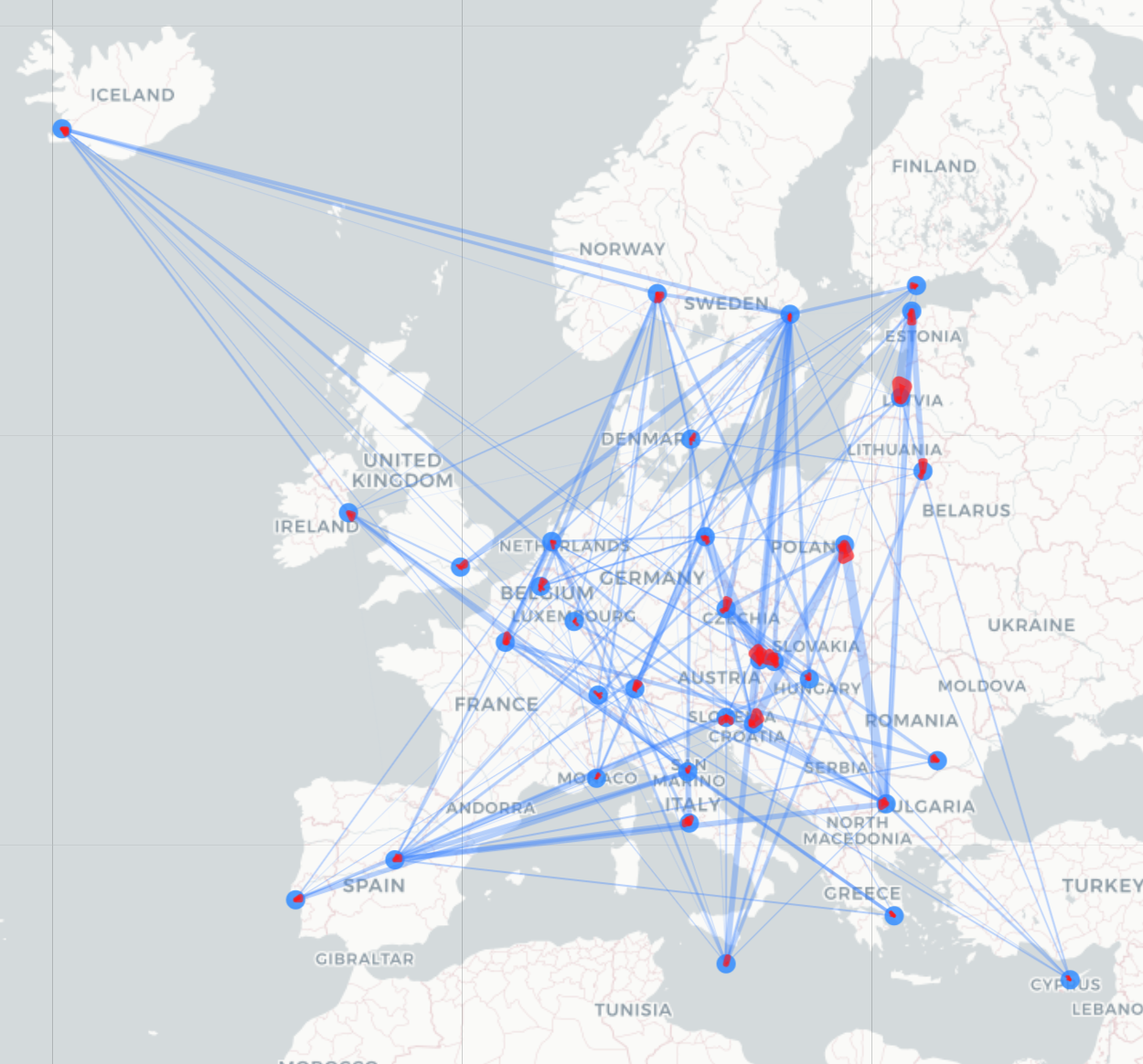}
    \caption{\textit{Inferred age dependency graph.} Non-zero values of $\widehat{\bm A}_{ij}$ are represented by weighted directed edges from country $j$ to country $i$, which take form of an arrow. Arrow head is colored in red to make the direction more visible. Thicker arrows correspond to larger weights. Blue circles around countries represent values of $\widehat{\bm A}_{ii}$. }
    \label{fig: age_pop}
\end{figure}

To apply Model \eqref{eq:wmar}, we firstly denote the distribution of age population, of country $i$, at year $T$, by $\bm \mu_{t}^i$, 
with $T = 1, ..., 40$ and $i = 1, ..., 34$. Note that the age considered by the data base goes through $0$ to $100$-plus. Thus we take the $100$-plus as $100$, and moreover scale down the age by $100$ to make the age distribution supported over $[0,1]$. Then we retrieve the quantile function $\bm F_{i,t}^{-1}$ of $\bm \mu_{t}^i$ from the population counts by ages of country $i$ recorded at year $t$, with numeric methods. 
\CR{We observed a trend in the raw distributional time series as illustrated in Figure \ref{fig:france_qt}. Since our method assumes that the time series is stationary under Definition \ref{def: stationarity hilbert space}. We propose to detrend the raw time series before model fitting. It consists in first estimating the trend, second subtracting it from the raw time series. In our context, the trend is a constant function from the domain of time to the Wasserstein space. An appropriate method to estimate the trend is Fréchet regression proposed in \cite{petersen2019frechet}, where the authors extended both classical global and local linear regression models to responses in a generic metric space. Therefore, for each component $i$, we regress the time series of distributions, $\bm \mu_{t}^i, t = 1, ..., 40$, on $t$ using the local Fréchet regression, and use the estimated regression function as the estimated trend, denoted by $m^i(t) \in \WR, t \in \R$. The adapted Fréchet regression to the Wasserstein space $\WR$ can be done only using quantile functions. We used the corresponding R package ``frechet'' for the implementation. To subtract the trend from $\bm \mu_{t}^i$, we rely again on quantile functions. Similarly to the proposed centering method, we compose the quantile functions of the raw data $\bm \mu_{t}^i$ with the ones of the trend as follows
\begin{equation}\label{eq: detrending}
  \bm F_{i,t}^{-1} \circ F_{i,\oplus}(t),
\end{equation}
where $ \bm F_{i,t}^{-1}$ is the quantile of $\bm\mu_{t}^i$, and $F_{i,\oplus}(t)$ is the cdf of $m^i(t)$. We display in Figure \ref{fig:detrending} an example of detrended series. 
}
Lastly, we use these detrended quantile functions to calculate the proposed estimator $\widehat{\bm A}$ in optimization problem\footnote{We use the same stopping criteria as in the simulation, while we apply the granularity of $0.002$. } \eqref{eq: vec_rep_cols}.
%In particular, we retrieve the quantile functions using  continuous functions which take $0$ on $p = 0$ and $1$ on $p = 1$ so as to be consistent with our assumptions (for details see function $\mathtt{generate\_qt\_fun}$ defined in script \textit{age\_pop.py} in the code related to this paper).
The complete execution time of model fitting takes around $78$ seconds. Figure \ref{fig: age_pop} shows a visualization of $\widehat{\bm A}$ as a directed weighted graph, overlapped on the geographical map. The graph in interactive form is available at \url{https://github.com/yiyej/Wasserstein_Multivariate_Autoregressive_Model}.

\begin{figure}[htbp]
    \centering

    % --- Ligne 1 : raw ---
    \begin{subfigure}[t]{0.48\linewidth}
        \centering
        \includegraphics[width=\linewidth]{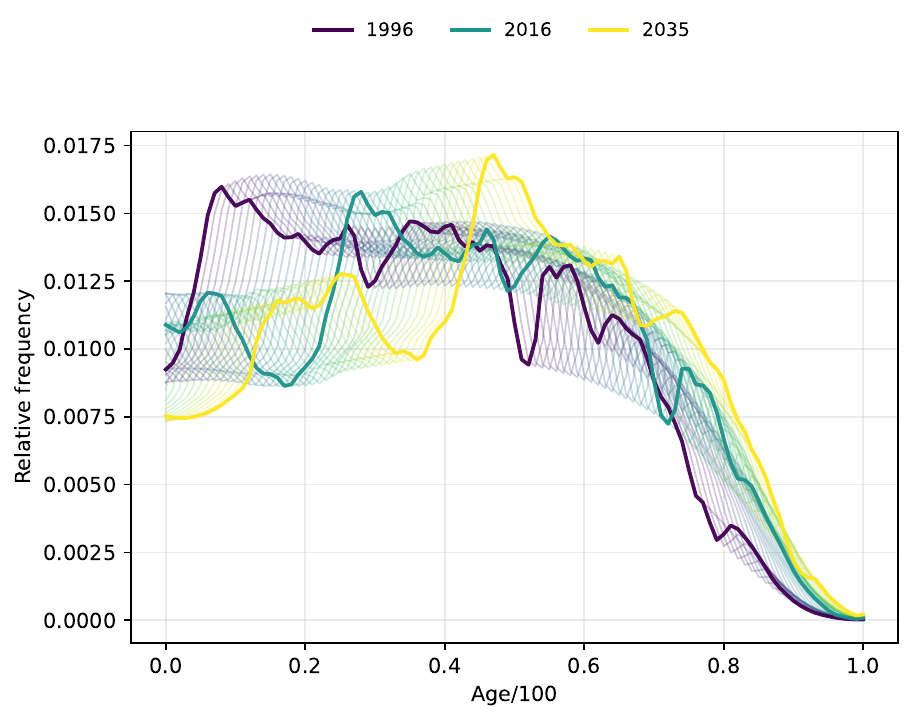}
        \caption{Estonia (raw)}
        \label{fig:estonia_raw}
    \end{subfigure}
    \hfill
    \begin{subfigure}[t]{0.48\linewidth}
        \centering
        \includegraphics[width=\linewidth]{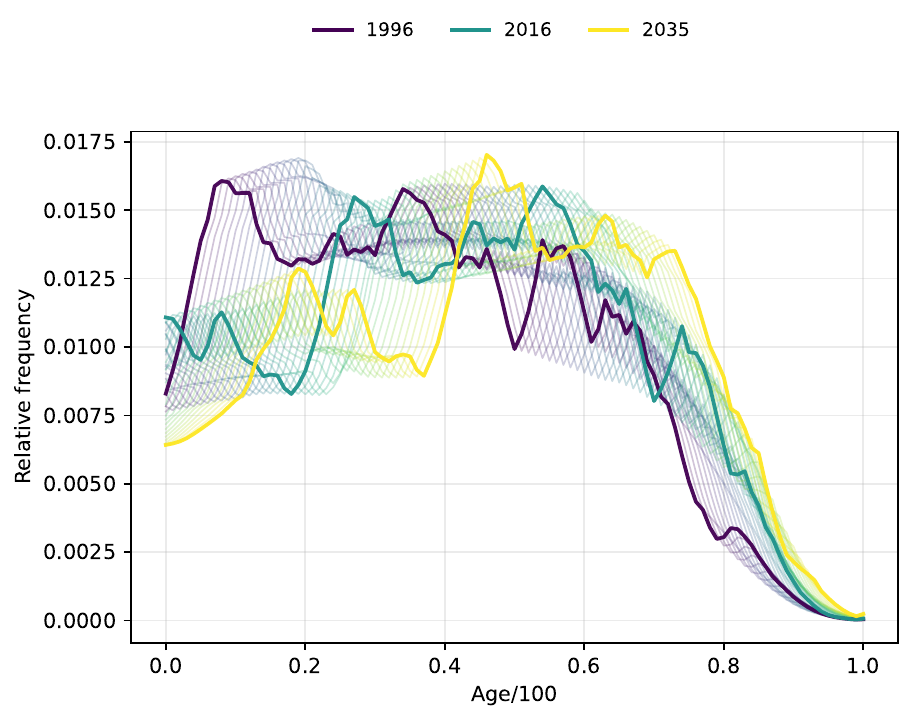}
        \caption{Latvia (raw)}
        \label{fig:latvia_raw}
    \end{subfigure}

    \vspace{0.7em}

    % --- Ligne 2 : detrended ---
    \begin{subfigure}[t]{0.48\linewidth}
        \centering
        \includegraphics[width=\linewidth]{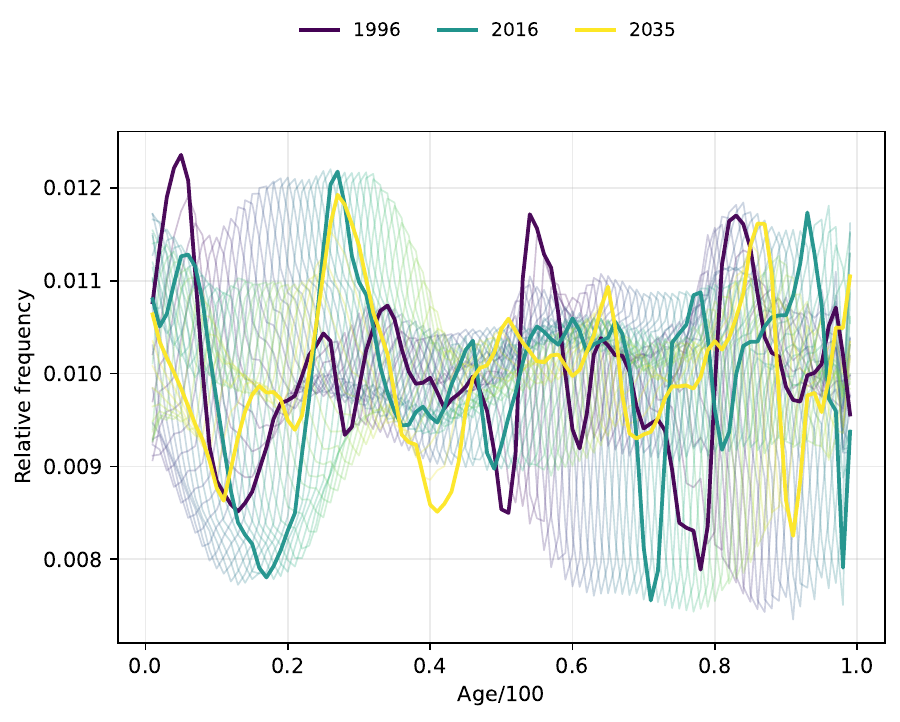}
        \caption{Estonia (detrended)}
        \label{fig:estonia_detr}
    \end{subfigure}
    \hfill
    \begin{subfigure}[t]{0.48\linewidth}
        \centering
        \includegraphics[width=\linewidth]{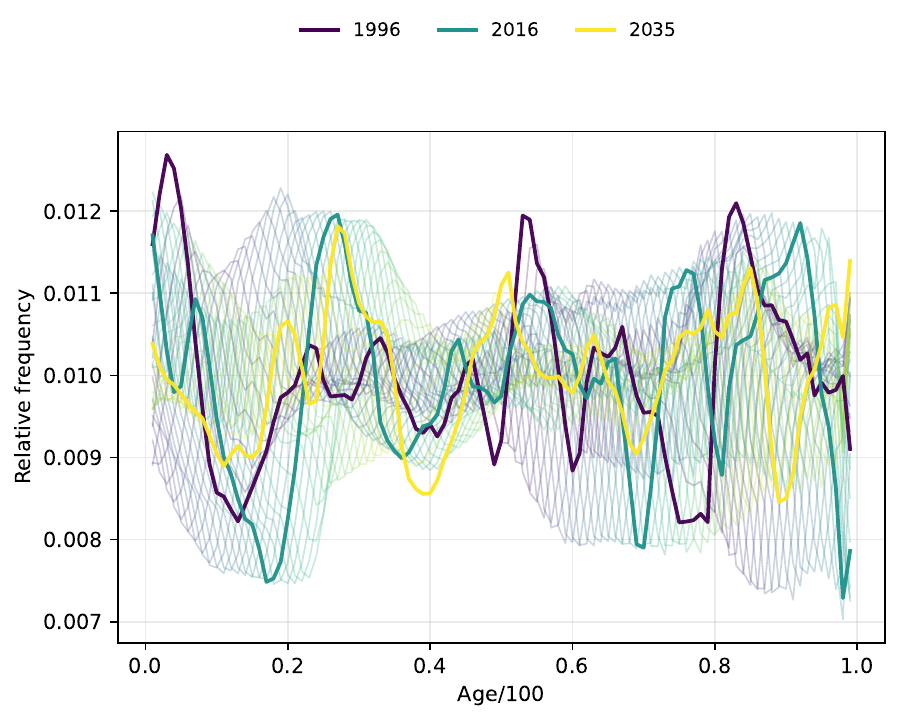}
        \caption{Latvia (detrended)}
        \label{fig:latvia_detr}
    \end{subfigure}

    \caption{\textit{Evolution of age structure from $1996$ to $2036$ (projected) of Estonia versus Latvia.}
    The relative frequencies corresponding to the detrended series are reconstructed from the detrended quantile functions $\bm F_{i,t}^{-1} \circ F_{i,\oplus}(t)$. }
    \label{fig:estonia_latvia}
\end{figure}

\begin{figure}[htbp]
    \centering

    % --- Ligne 1 : raw ---
    \begin{subfigure}[t]{0.48\linewidth}
        \centering
        \includegraphics[width=\linewidth]{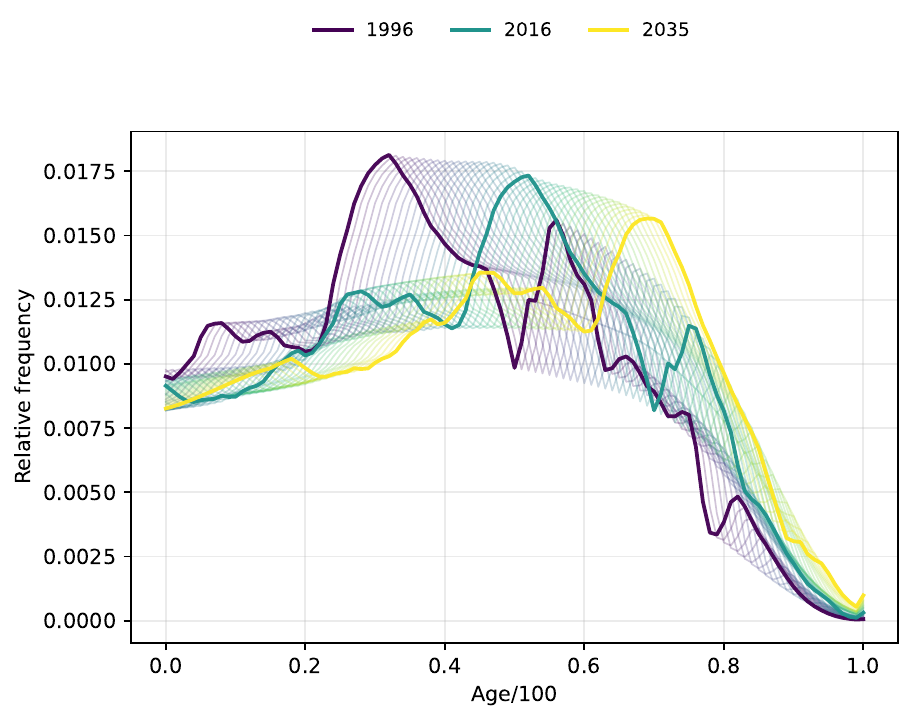}
        \caption{Germany (raw)}
        \label{fig:germany_raw}
    \end{subfigure}
    \hfill
    \begin{subfigure}[t]{0.48\linewidth}
        \centering
        \includegraphics[width=\linewidth]{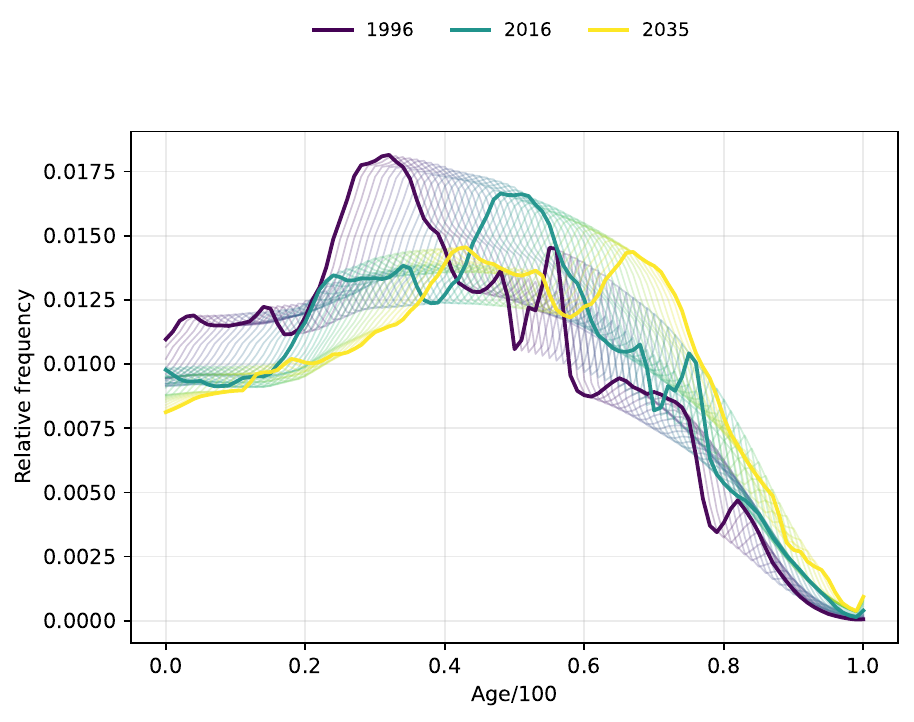}
        \caption{Austria (raw)}
        \label{fig:austria_raw}
    \end{subfigure}

    \vspace{0.7em}

    % --- Ligne 2 : detrended ---
    \begin{subfigure}[t]{0.48\linewidth}
        \centering
        \includegraphics[width=\linewidth]{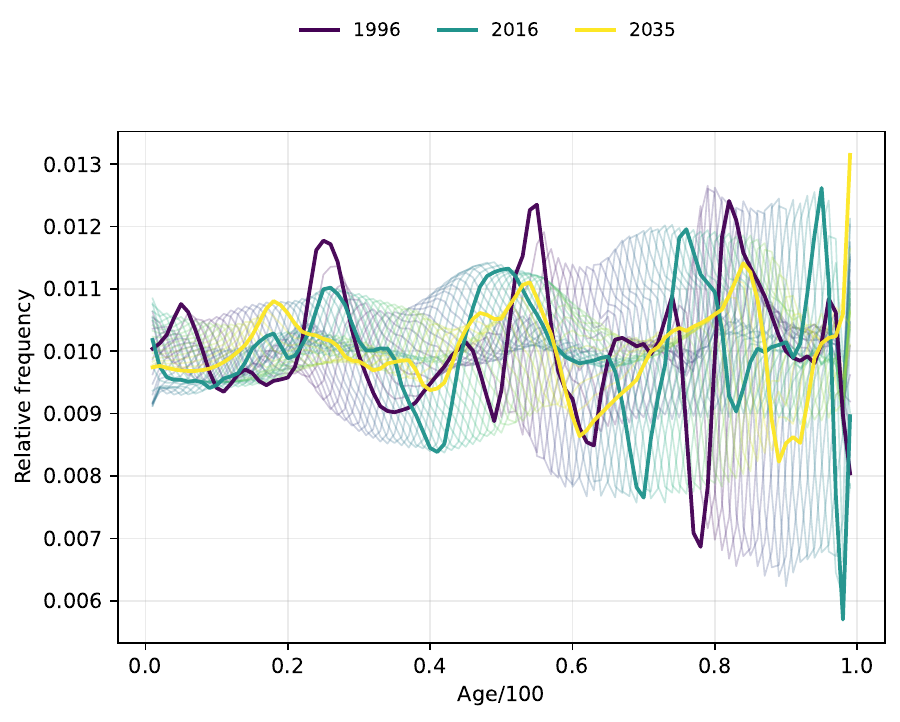}
        \caption{Germany (detrended)}
        \label{fig:germany_detr}
    \end{subfigure}
    \hfill
    \begin{subfigure}[t]{0.48\linewidth}
        \centering
        \includegraphics[width=\linewidth]{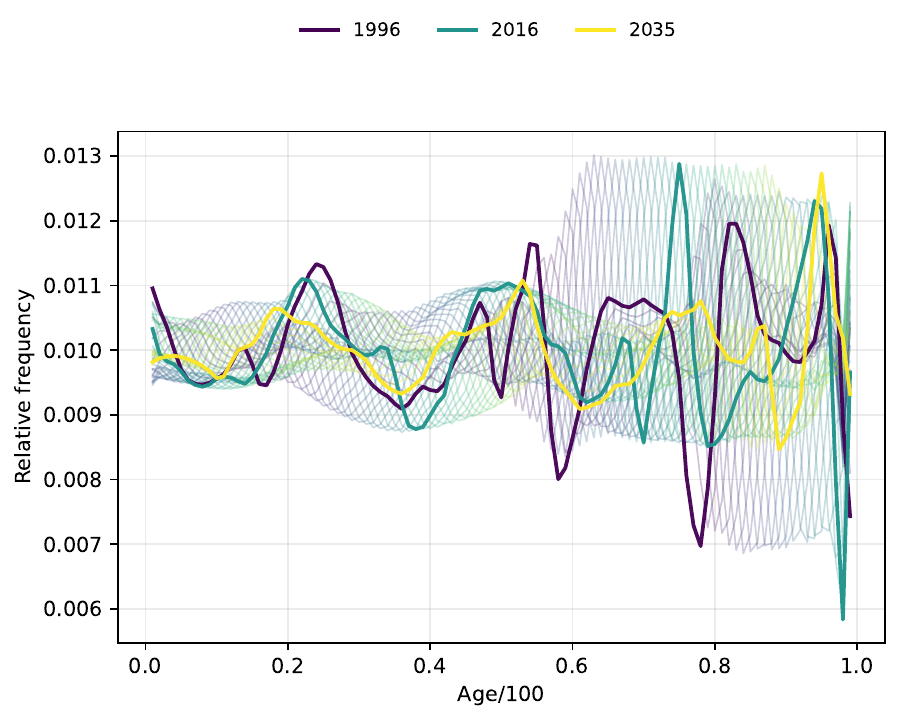}
        \caption{Austria (detrended)}
        \label{fig:austria_detr}
    \end{subfigure}

    \caption{\textit{Evolution of age structure from $1996$ to $2036$ of Germany versus Austria.}}
    \label{fig:germany_austria}
\end{figure}

Firstly, some countries are not connected such as Iceland and Portugal. This means their corresponding values of $\widehat{\bm A}_{ij}$ are zero. This is because Assumption \ref{assump: B1+} has the same effect as Lasso regularization in optimization \eqref{eq: op L2}. Thus the estimator $\widehat{\bm A}$ is naturally sparse, which helps us to identify significant dependencies between series more easily. 
%Secondly, for all countries, the weight of $\widehat{\bm A}_{ii}$ dominates the weights of $\widehat{\bm A}_{ij}, \, i \neq j$.
%%which are bounded by $0 \leq \sum_{j=1}^{34} A_{ij} \leq 1$. 
%This is because the age structure of a country does not change much between two consecutive years. On the other hand, this also implies the age structure differs largely across countries. Thirdly, the model indicates "significant" cross-country links, which can motivate further demographic studies.  
%in the sense that they are not suppressed to zero by the $\ell_1$ regularization $0 \leq \sum_{j=1}^{34} A_{ij} \leq 1$. 
We investigate the first two largest cross-country weights, which are respectively: Estonia $\rightarrow$ Latvia, and Germany $\rightarrow$ Austria. To justify the inferred edges, we plot the time series of age distributions of these four countries in Figures \ref{fig:estonia_latvia} and \ref{fig:germany_austria}. For clearer visualization, we plot the time series of distributions in terms of relative frequencies rather than quantile functions, since the latter are so densely packed that differences across distributions are difficult to discern, as shown in Figure \ref{fig:detrending}.
% here we only call age structure but not distribution in the context of Figures. Since the empirical distribution should be discontinuous/ discrete points. Here to facilitate the visualization, we connect these empirical probability, however, the connecting segments between relative frequencies do not represent probability from estimation point of view.  

\begin{figure}[htbp]
    \centering

    % --- Ligne 1 : raw ---
    \begin{subfigure}[t]{0.48\linewidth}
        \centering
        \includegraphics[width=\linewidth]{images/France_rainbow.pdf}
        \caption{France (raw)}
        \label{fig:France_raw}
    \end{subfigure}
    \hfill
    \begin{subfigure}[t]{0.48\linewidth}
        \centering
        \includegraphics[width=\linewidth]{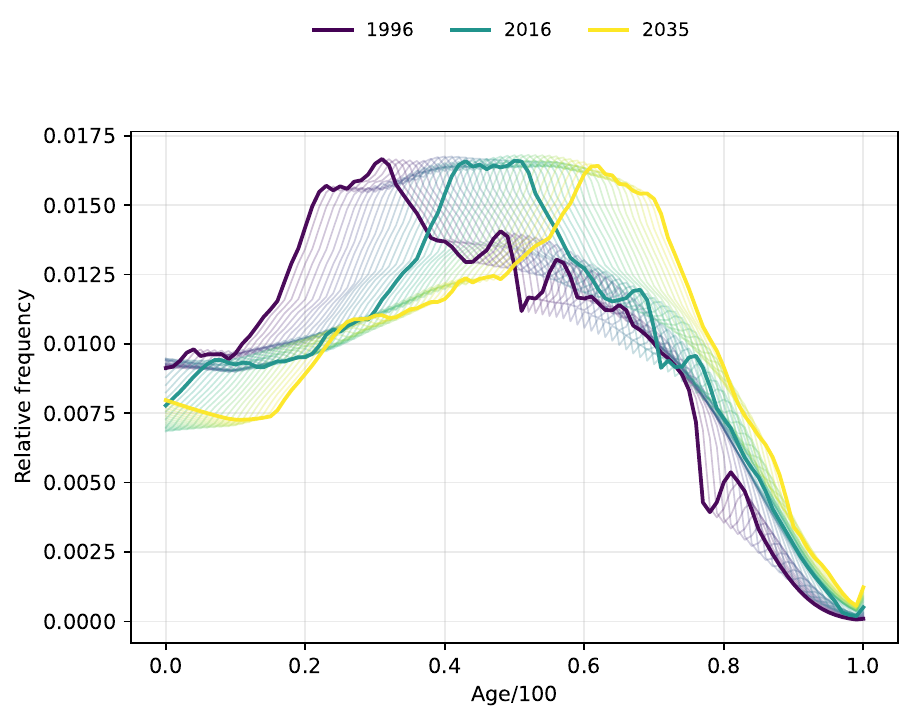}
        \caption{Italy (raw)}
        \label{fig:Italy_raw}
    \end{subfigure}

    \vspace{0.7em}

    % --- Ligne 2 : detrended ---
    \begin{subfigure}[t]{0.48\linewidth}
        \centering
        \includegraphics[width=\linewidth]{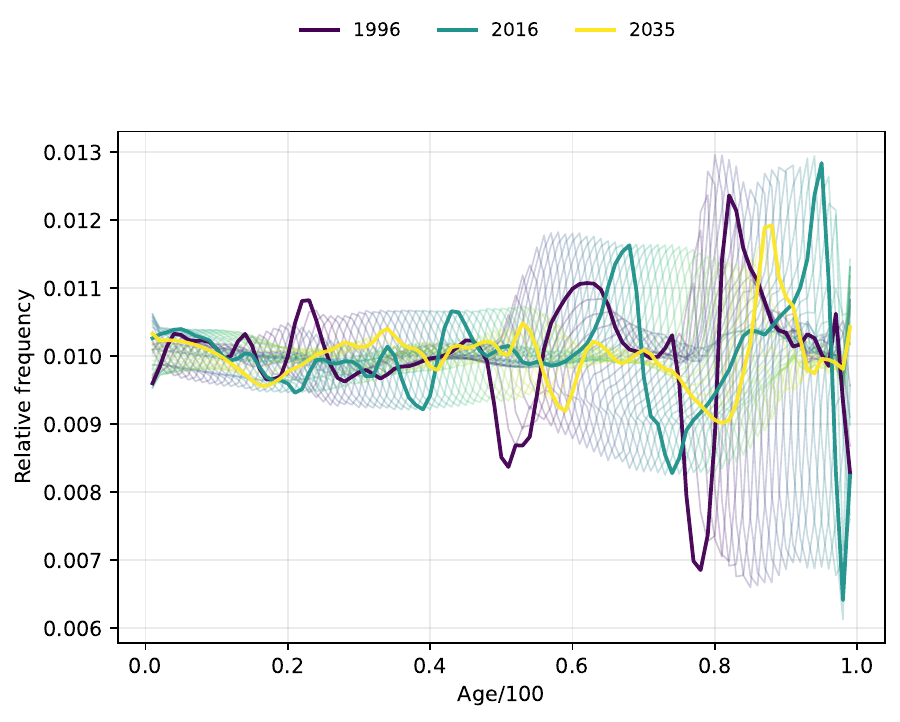}
        \caption{France (detrended)}
        \label{fig:France_detr}
    \end{subfigure}
    \hfill
    \begin{subfigure}[t]{0.48\linewidth}
        \centering
        \includegraphics[width=\linewidth]{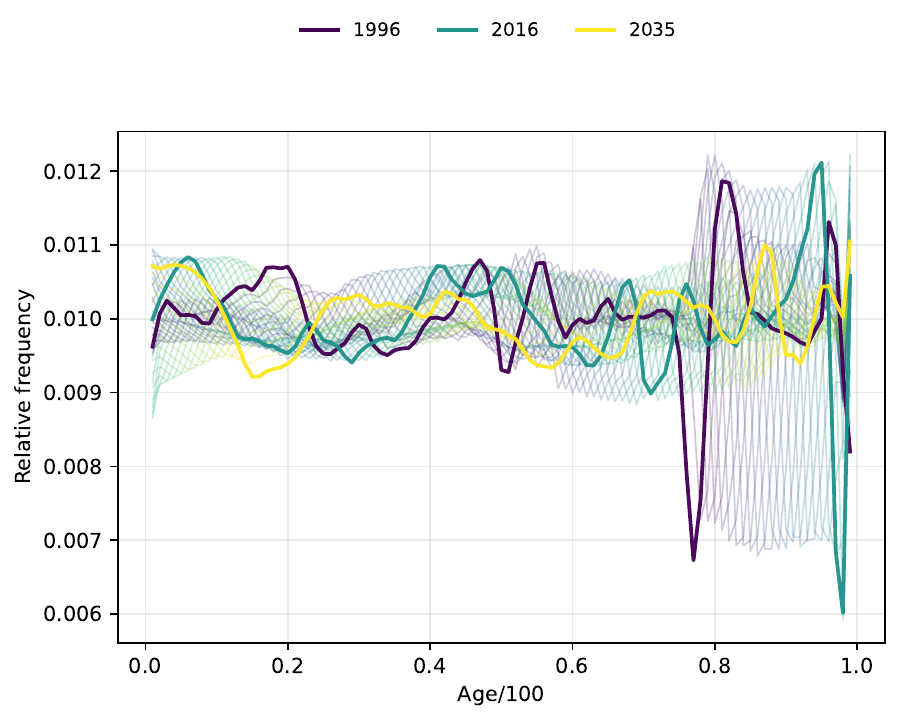}
        \caption{Italy (detrended)}
        \label{fig:Italy_detr}
    \end{subfigure}

    \caption{\textit{Evolution of age structure from $1996$ to $2036$ (projected) of France versus Italy.}}
    \label{fig: france italy}
\end{figure}

\CR{Recall that the links are inferred from the detrended series, thus similarity between the detrended series of linked countries can be seen. By contrast, the detrended series between unlinked countries are very different,  for example Estonia in Figure \ref{fig:estonia_detr} versus Germany in Figure \ref{fig:germany_detr}. We also plotted the raw series, which are likewise very similar among linked countries. This suggests that these countries' population structures are closely associated: they share a common long-term transition pattern (trend), and short term shocks (detrended series) of the outgoing countries propagate rapidly to their incoming counterparts. }
Indeed, these two linked pairs consist both of bordering countries. To furthermore justify our model, we check neighbouring countries which are not linked, for example France and Italy, in Figure \ref{fig: france italy}. Their age structures are shown very different.

Lastly, in Table \ref{tbl: top weights}, we provide the first $5$ edges of the largest weights. 
The proposed model reports patterns consistent with the geographical and demographic facts, but is also able to identify new features, which can potentially motivate the follow-up studies in the related domains. These observations show the credibility and usefulness of the proposed model.

\begin{table}[htbp]
\begin{center}
\caption{\textit{Top $5$ edges with the largest weights excluding all the self-loops}}
\vspace{0.1in}
\label{tbl: top weights}
\begin{tabular}{||c c c||}
 \hline
  & From & To\\ [0.5ex] 
 \hline\hline
 1 & Estonia & Latvia \\ 
 \hline
 2 & Germany&
Austria\\
 \hline
 3 & Bulgaria&
Poland\\
 \hline
 4 & Czech Republic&
Slovakia\\
 \hline
 5 & Poland&
Croatia\\ [1ex] 
 \hline
\end{tabular}
\end{center}
\end{table}

\subsection{Bike-sharing network in Paris}\label{sec: real data set bike sharing}
\begin{figure}
    \centering
    \begin{subfigure}[t]{\textwidth}
        \centering
    \includegraphics[width=0.56\linewidth]{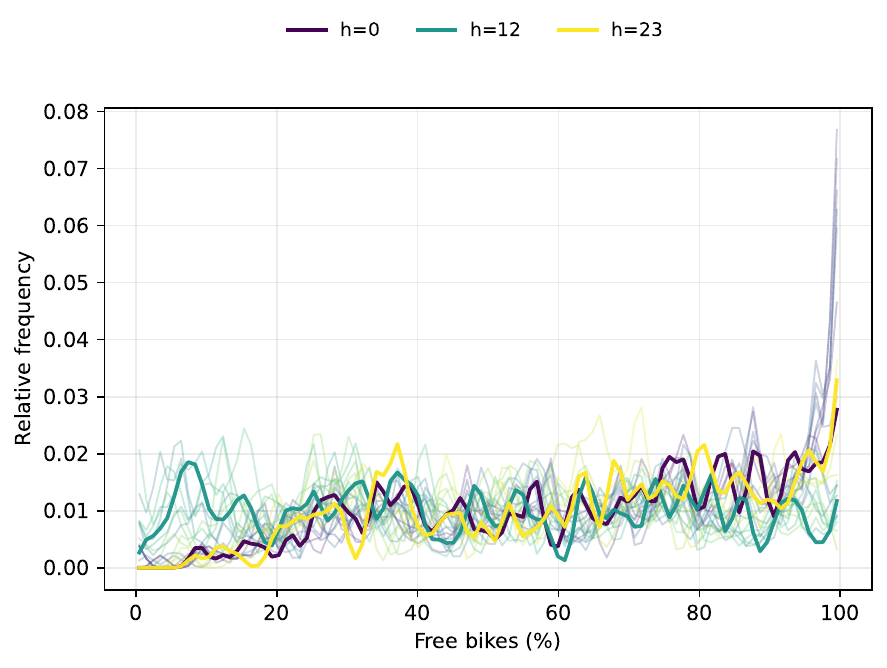}
        \caption{Raw series in relative frequency}
    \end{subfigure}
    \begin{subfigure}[t]{0.48\textwidth}
        \centering
        \includegraphics[width=\linewidth]{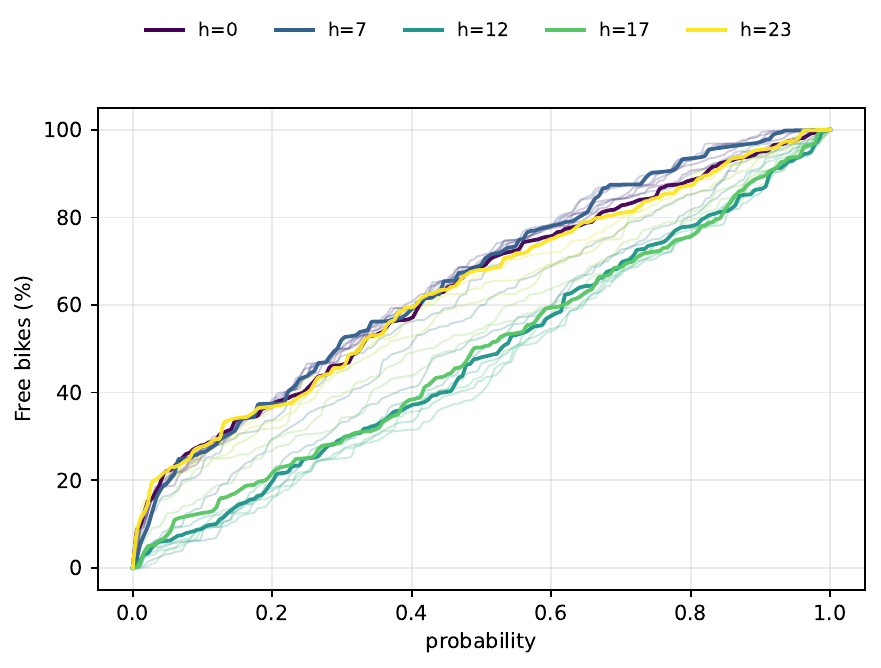}
        \caption{Raw series in quantile functions $\bm F_{i,t}^{-1}$}
    \end{subfigure}
    \begin{subfigure}[t]{0.48\textwidth}
        \centering
        \includegraphics[width=\linewidth]{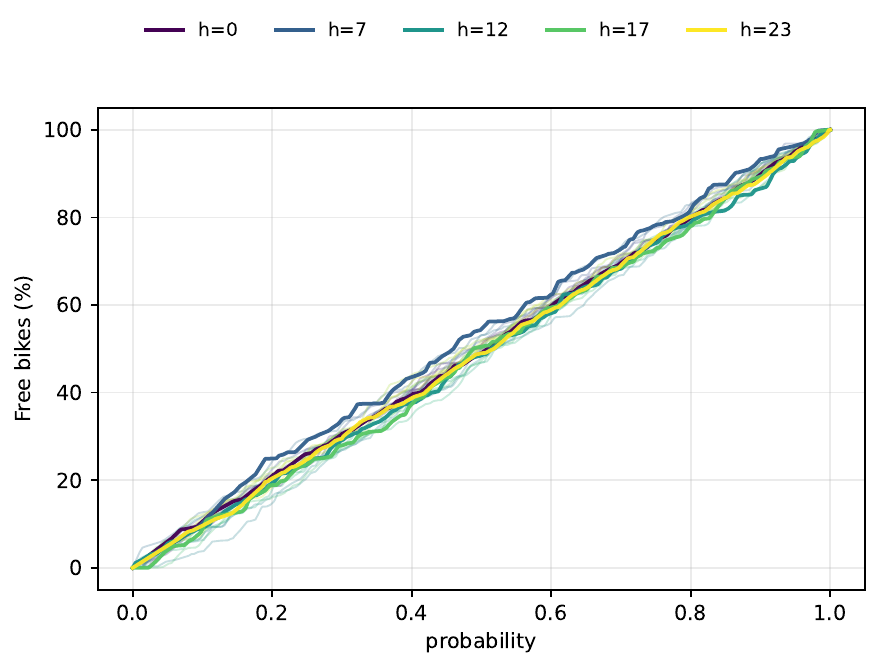}
        \caption{Detrended series $\bm F_{i,t}^{-1} \circ F_{i,\oplus}(t)$}
    \end{subfigure}
    \caption{\textit{Retrieved series of distributions of Station digeon-saint-mande.} }
    \label{fig: digeon-saint-mande}
\end{figure}

\begin{figure}
    \centering
    \includegraphics[width=0.6\linewidth]{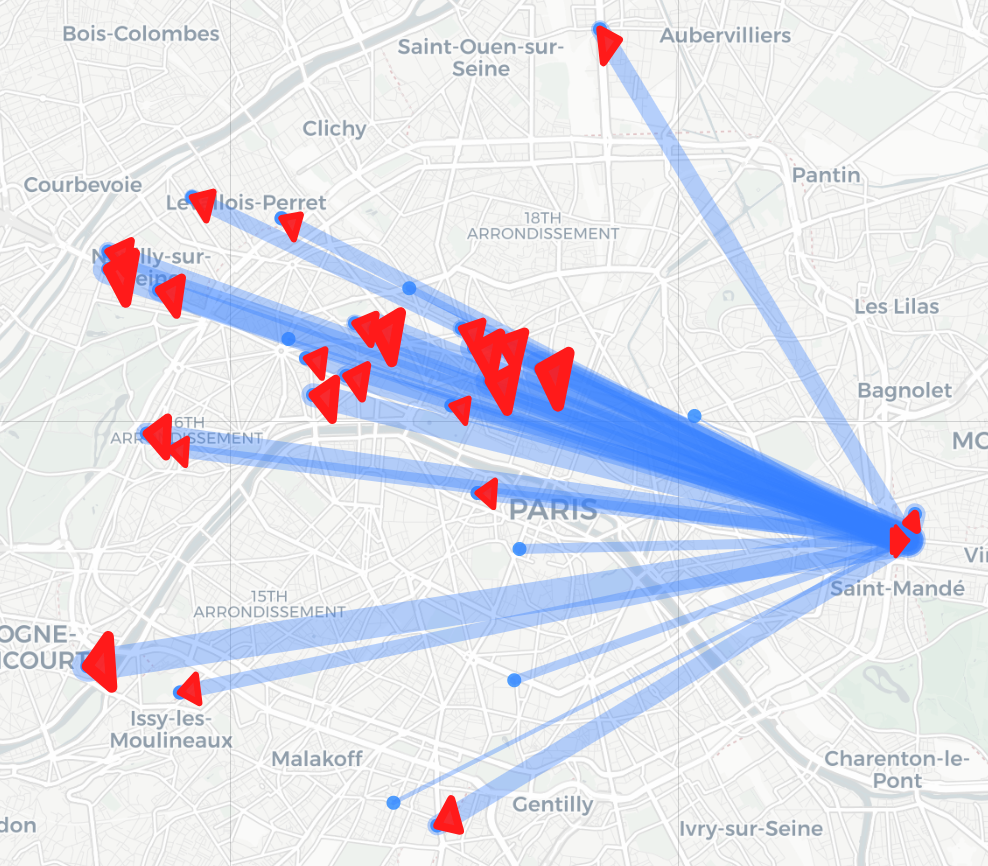}
    \caption{\textit{Outward hub: Station lagny-saint-mande, its top 25 thickest outward edges and its top 25 thickest inward edges.} }
    \label{fig: bike-sharing-Paris_subgraph1}
\end{figure}

\begin{figure}
    \centering
    \includegraphics[width=0.65\linewidth]{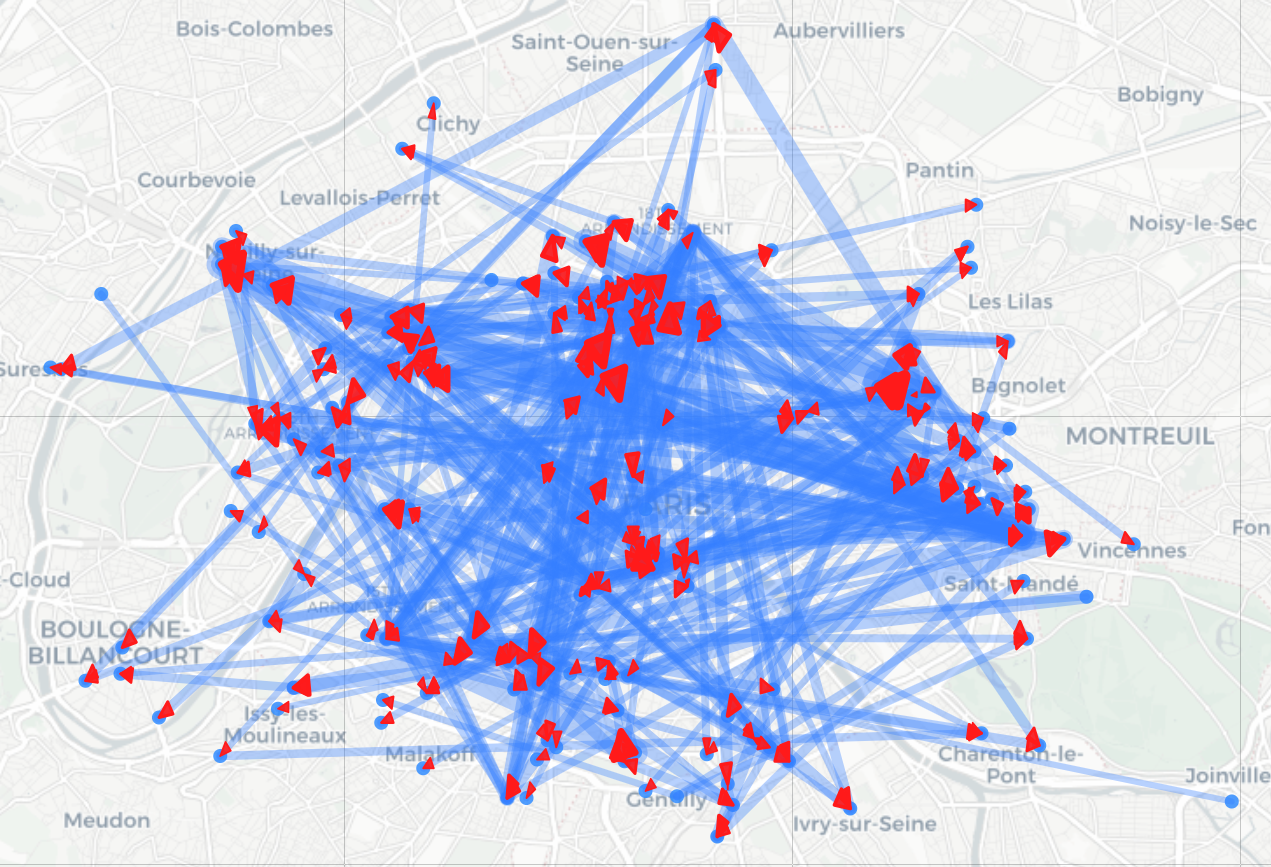}
    \caption{\textit{$21$-core.} For better visualization, only the top $10\%$ thickest edges in the subgraph are shown.}
    \label{fig: bike-sharing-Paris_subgraph2}
\end{figure}

To furthermore strength the credibility of the proposed model, we test it on another real data set, which is the bike-sharing data set of Paris from \citep{jiang2020sensor}. The data set records the ratio of available bikes of $274$ stations, observed over $4417$ consecutive hours. We are interested in the temporal evolution of bike availability at these stations, and would like to identify how the stations relate mutually in their evolution using the proposed method. To this end, we firstly represent the data by taking into account the distribution aspect. \CB{We introduce the temporal variable $t$ which represents hours in a day. Accordingly, the distribution $\bm \mu_{t}^i$ considered by Model \eqref{eq:wmar} represents the distribution of bike availability of the station $i$, at hour $t$ in a day, with $t = 1, ..., 24$ and $i = 1, ..., 274$. Thus the quantile function $\bm F^{-1}_{i,t}$ of distribution $\bm \mu_{t}^i$ can be retrieved from data points of station $i$, hour $t$ from different days. Note that since these observations are distanced from each other in time, they can be considered approximately as being independent samples.  An example of the retrieved distributions is given in Figure \ref{fig: digeon-saint-mande}.} \CR{Similarly as before, we observed trend in the raw time series. Thus we performed the same detrending method (Equation \eqref{eq: detrending}), the detrended series shown in Figure \ref{fig: digeon-saint-mande} as well.}
%We use the similar method as for the age distribution data to retrieve the valid quantile functions, however the implementation is different due to the availability of different quantity types (for details see function $\mathtt{generate\_qt\_fun}$ defined in script \textit{bike\_net.py} in the code related to this paper).
%For more comments on this point, we refer to Remark \ref{rem: how to define distribution of bike sharing}. 
We then fit Model\footnote{We use the same stopping criteria as previously, and we apply the granularity of $0.002$.} \eqref{eq: wmar_F} on the detrended series $\bm F_{i,t}^{-1} \circ F_{i,\oplus}(t), \, T = 1, ..., 24, \, i = 1, ..., 274$. The complete execution time of model fitting takes around $20$ minutes. We now demonstrate the visualization of $\widehat{\bm A}_{ij}$ on the map of Paris, represented by a directed weighted graph. Since the edges of the complete graph will be densely (even though the graph is very sparse with $274$ nodes and $4557$ edges) located in the plot when fitting the graph to the paper size. For better visual effects, we zoomed into two parts of the graph in Figure \ref{fig: bike-sharing-Paris_subgraph1} and Figure \ref{fig: bike-sharing-Paris_subgraph2} respectively. The complete graph is available in interactive form at \url{https://github.com/yiyej/Wasserstein_Multivariate_Autoregressive_Model}. 

\begin{figure}
    \centering
    \begin{subfigure}[t]{0.48\textwidth}
        \centering
        \includegraphics[width=\linewidth]{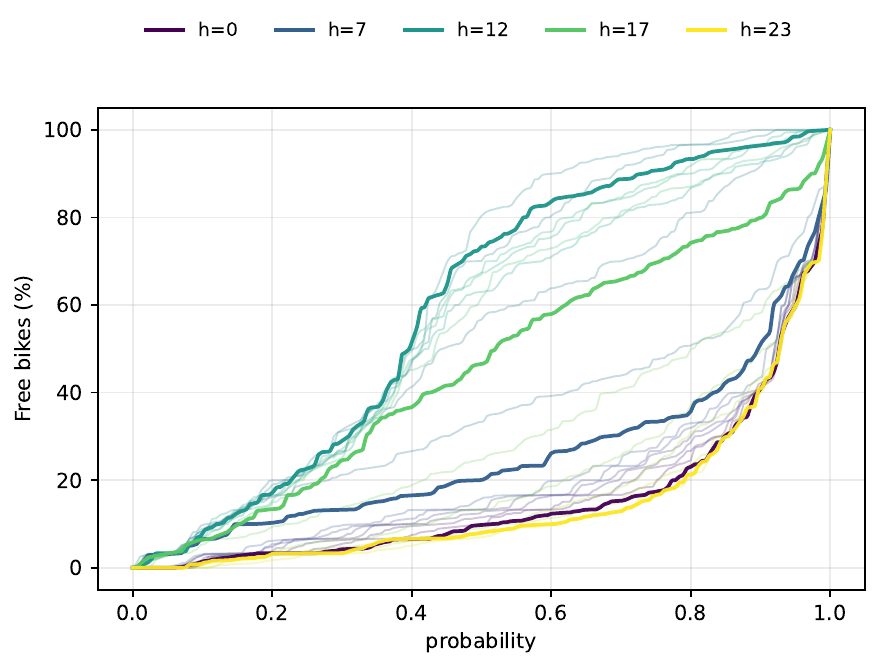}
        \caption{Raw series in quantile functions $\bm F_{i,t}^{-1}$}
    \end{subfigure}
    \begin{subfigure}[t]{0.48\textwidth}
        \centering
        \includegraphics[width=\linewidth]{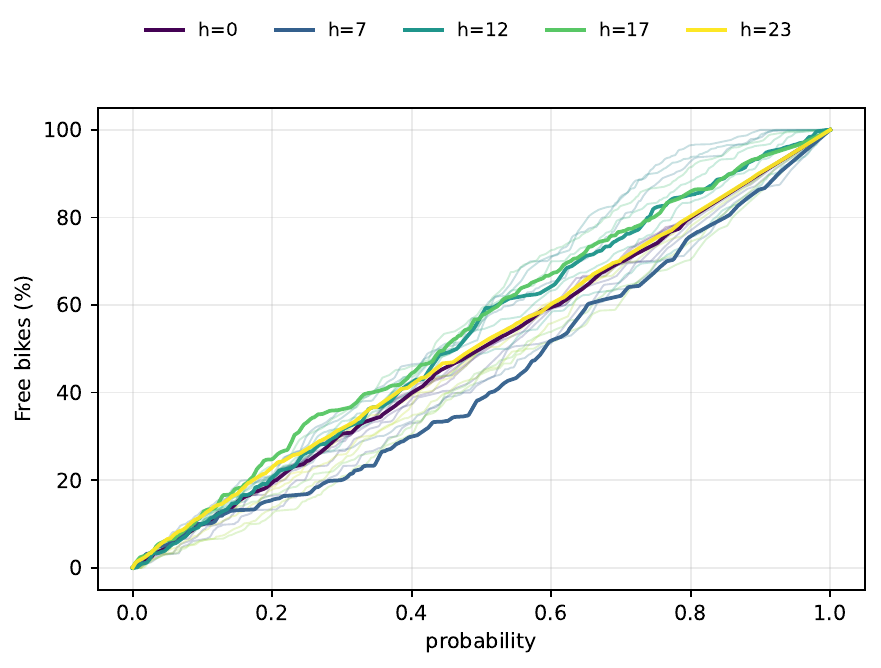}
        \caption{Detrended series $\bm F_{i,t}^{-1} \circ F_{i,\oplus}(t)$}
    \end{subfigure}
    \begin{subfigure}[t]{0.48\textwidth}
        \centering
        \includegraphics[width=\linewidth]{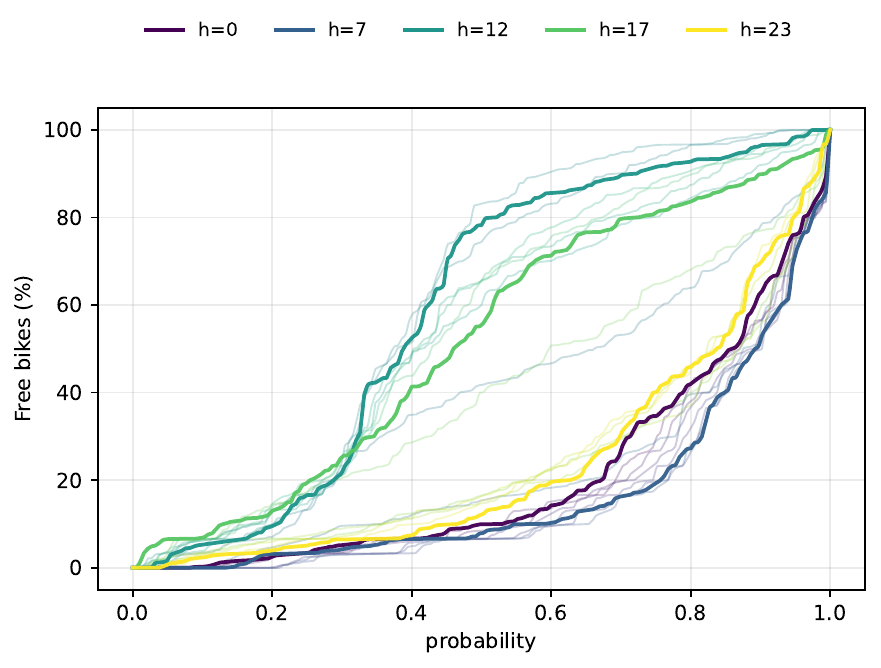}
        \caption{Raw series in quantile functions $\bm F_{i,t}^{-1}$}
    \end{subfigure}
    \begin{subfigure}[t]{0.48\textwidth}
        \centering
        \includegraphics[width=\linewidth]{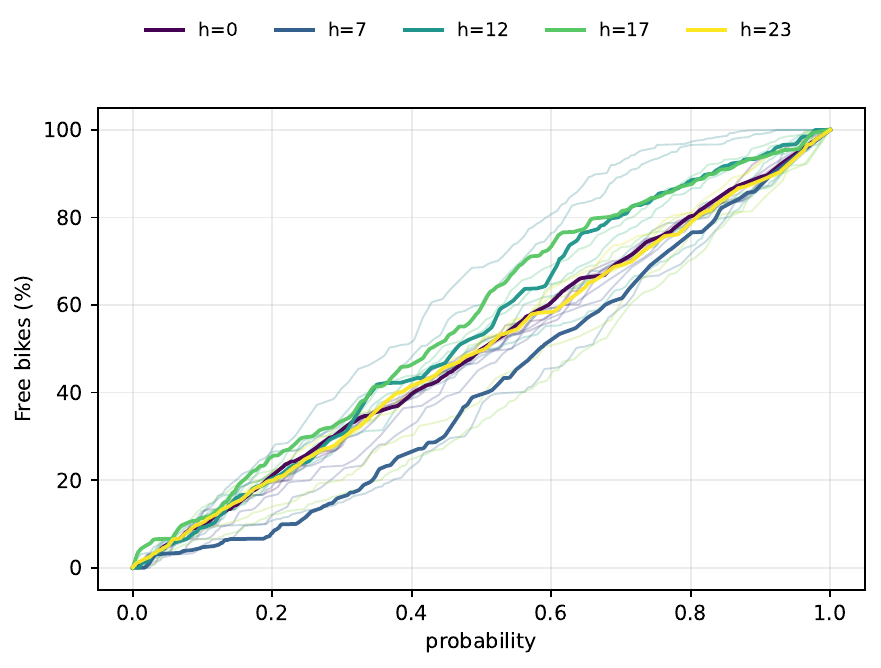}
        \caption{Detrended series $\bm F_{i,t}^{-1} \circ F_{i,\oplus}(t)$}
    \end{subfigure}
    \caption{\textit{Time series of Stations lagny-saint-mande (top) and rougemont (bottom).} }
    \label{fig: digeon-saint-mande rougemont}
\end{figure}

\CR{Figure \ref{fig: bike-sharing-Paris_subgraph1} reports the outward hub, which has the maximal sum of outward weights, as well as its top links. The hub station \textit{lagny-saint-mande} has strong links with the stations along the flow from itself in southeast to the station \textit{de-gaulle-3-neuilly} in northwest. This flow actually goes along the main itinerary of Metro $1$ and RER A in Paris, which starts around \textit{Saint-Mandé}. We plot the time series of the hub station \textit{lagny-saint-mande} and its nearest neighbour \textit{rougemont} in Figure \ref{fig: digeon-saint-mande rougemont}. We can see that the two time series are very similar especially compared to the series of Station \textit{digeon-saint-mande} displayed in Figure \ref{fig: digeon-saint-mande}. By contrast, geographically, \textit{digeon-saint-mande} is very closed to \textit{lagny-saint-mande}. This means that our model is able to discover meaningful patterns from data other than the natural ones induced by a  geographical map.}

\CR{Additionally, we report the $21$-core of the graph in Figure \ref{fig: bike-sharing-Paris_subgraph2}. The $k$-core of a graph is its biggest subgraph, where each node has at least $k$ edges with the nodes in the same subgraph. Thus it can extract the main interactions of the graph. From Figure \ref{fig: bike-sharing-Paris_subgraph2}, we can see for example two strands of strong edges that cross Paris, one has been reported in Figure \ref{fig: bike-sharing-Paris_subgraph1} that goes from \textit{Saint-Mandé} in the southeast to the northwest. The other one is approximately perpendicular to the \textit{Saint-Mandé} strand, which starts from \textit{Gare du nord} in the south to the north. Along these strands, changes in use of bikes at outward stations propagate to the inward stations such that their detrended time series resemble. Recall that the graph was inferred from the detrended data. In addition, the shown $21$-core indicates the busiest stations and areas in Paris.}

%We can also notice other interesting cross-region features from the two subgraphs. For example, it seems the dynamic of the railway station \textit{Gare de du Nord} () is unpredictable by other stations in Paris, nevertheless it is very useful  in predicting others, especially the other railway station \textit{Gare Montparnasse} in  (bottom of Figure \ref{fig: bike-sharing-Paris_subgraph1}).

% \begin{remark}\label{rem: how to define distribution of bike sharing}
% Note that another way to represent the data set by distributions is to consider every $K$ consecutive observations as the i.i.d. samples of a distribution. Thus $T = 1, 2, ..., 185/K$ represents the ongoing time in hours. However, the consecutive observations are highly correlated. Moreover, the distribution retrieved in this manner does not clearly represent a random variable. Thus we do not go further with this data representation method.
% \end{remark}

\section{Conclusion}
In this paper, we extend the standard VAR models to distributional multivariate AR models, which provides an approach to model a collection of time series of univariate probability distributions, and to represent their dependency structure by a directed weighted graph at the same time. Especially, the proposed data centering method for random measures allows the development of other AR and regressive models with multiple predictors. Moreover, our empirical studies on real data sets demonstrate that, the proposed models equipped with the distributional data representation are efficient tools for analyzing and understanding the spatial-temporal data. 
%In particular, this paper provides a class of multivariate AR models for distributional time series that favors sparse estimation of  AR coefficients  which is beneficial for graph learning.

\section*{Declaration of Competing Interest}
The authors declare no competing interests.

\bibliographystyle{plainnat}

\bibliography{main}
\end{document}